# Multi-Strategy Improved Snake Optimizer Accelerated CNN-LSTM-Attention-Adaboost for Trajectory Prediction

Shiyang Li [1]

[1] Civil Aviation University of China

**Abstract:** 为了解决中长期四维（4D）轨迹预测模型的局限性，本文提出一种融合多策略改进蛇群优化算法（SO）的 CNN-LSTM-attention-adaboost 混合神经网络模型。该模型应用 adaboost 算法思想划分出多个弱学习器，每个子模型利用 CNN 提取空间特征，利用 LSTM 来捕捉时序特征，并利用注意力机制全面捕捉全局特征。由多个子模型组合而成的强学习器模型再通过 SO 所模拟的自然选择行为模式来优化预测模型的超参数。本研究基于西安到天津的真实 ADS-B 数据，进行了多项优化器的对比实验和消融研究，并开展了全面的测试评估分析。结果表明，SO-CLA-adaboost 在处理大规模高维轨迹数据方面优于粒子群、鲸鱼、灰狼等传统优化器。此外，全策略协同改进 SO 算法的引入，使模型的预测精度提高了 39.89%。

# 一、引言

随着全球航空运输量的持续增长，空域资源日益紧张，航空器运行安全面临新的挑战。国际航空运输协会(IATA)数据显示，2023 年全球航空客运量已恢复至疫情前水平的 95%，预计 2040 年航空器日均航班量将突破 20 万架次。《中国民航 2024》公布的统计数据显示，中国民航运输业总周转量、旅客运输量、货物运输量等各项运输指标均呈快速上升趋势。[1] 在此背景下，精确的航空器航迹预测技术成为提升空域容量、保障飞行安全的关键支撑。现有的空中交通管理系统对轨迹预测的准确性和实时性提出了更高的要求，而传统的预测方法在复杂气象条件和空管指令变化时表现欠佳，这促使我们探索融合多源数据的智能预测新范式。[2]

航迹预测技术历经多年发展，主要形成了基于运动学模型、基于状态估计以及基于机器学习与数据驱动这三大类方法。

基于运动学模型的航迹预测方法，依据物理规律选取航空器的飞行性能参数，如速度、航向、爬升率等，对飞行阶段进行划分，借助运动学微分方程实现短期航迹预测。这类模型对风速、天气等外界不确定因素的响应能力较弱，难以适应飞行过程中的突发动态变化，导致在长时间航迹预测中精度欠佳。Kaneshige J 及其团队基于航空器动力学模型，将航空器分解为水平与高度剖面，对比各阶段航空器运行时间、机建燃油消耗量等指标，对比有无轨迹的的预测结果。结果表明，基于轨迹的预测精度高于无轨迹的预测。[3]Lin Y 等人提出基

于位置间相对运动的飞行计划轨迹预测算法，通过分析相对运动关系进行轨迹预测。[4]G Avanzini 利用 Frenet 坐标系相关算法进行轨迹预测，具体是通过对 Frenet 坐标系下的轨迹进行处理和分析来实现。[5]

基于状态估计法的航迹预测方法将航迹预测问题视为线性混合系统的状态估计问题。这类模型的时间复杂度呈指数级增长，在处理大规模长尺度航迹数据时，难以满足实时性要求。卡尔曼滤波器、交互式多模型（IMM）也被研究和应用于航迹预测。中国民航大学的吕波与王超利用改进的扩展卡尔曼滤波技术对 4D 航迹进行预测，通过引入航空器运动的六自由度模型，设立风场扰动函数，并构建状态变 量和输入变量，建立了非线性控制系统。[6] J. L. Yepes 等人提出新算法，通过推断飞机意图并结合状态估计和飞行模式估计进行轨迹预测。[7]Xi L 等人基于 ADS - B 技术进行飞机轨迹预测算法的仿真研究，分析相关算法在 ADS - B 数据下的轨迹预测性能。[8]I Hwang 等人将混合系统状态估计方法应用于飞机跟踪，通过对相关状态的估计实现对飞机轨迹的预测。[9]

基于机器学习与数据驱动的航迹预测方法通过对大规模航迹数据进行特征挖掘，构建复杂模型来实现对未来轨迹的预测。

神经网络作为机器学习领域的关键技术，通过模拟生物神经元的信息传递和网络结构构建数学模型，借助大量神经元的相互连接来模拟信息处理过程。深度学习则在神经网络的基础上，利用多层非线性感知器对输入数据进行多维特征提取，从而实现对复杂问题的建模与求解。Zhiyuan Shi 等人提出基于约束长短期记忆网络（cLSTM）的 4D 飞行轨迹预测模型，通过构建爬升、巡航和下降 / 进近阶段的约束，结合 LSTM 网络实现高精度轨迹预测。[10] Ping Han 等人提出结合 K-means 聚类和门控循环单元（GRU）神经网络的在线学习模型，通过 K-means 聚类处理轨迹数据，利用 GRU 神经网络进行预测。[11] Jin Huang 和 Weijie Ding 提出基于贝叶斯优化的时间卷积网络-双向门控循环单元（DSA-TCN-BiGRU）混合神经网络模型，利用贝叶斯算法优化超参数，提高轨迹预测性能。[12]P Jia 等人构建基于注意力机制和长短期记忆网络（Attention-LSTM）的飞机 4D 轨迹预测模型，利用注意力机制增强 LSTM 对关键信息的捕捉能力以实现精准预测。[13]Zhonghang Fan 等人提出基于残差递归神经网络（RESRNN）的轨迹预测方法，结合 RNN 和残差神经网络进行轨迹预测。[14]

近年来，学术界从动物种群捕食行为中获取灵感，引入了一系列仿生智能算法，如灰狼算法、鲸鱼算法、冠猪优化算法等，为神经网络参数优化提供了全新的思路与方法，有望进一步提升深度学习模型的性能与应用效果。针对现有航迹预测方法在处理时空动态特征、非线性问题以及模型超参数优化方面存在的局限性，本研究提出一种基于多策略改进的蛇群优化算法的 CNN - LSTM - attention-adaboost 混合神经网络模型框架，旨在通过改进后的蛇群算法提高模型超参数选择效率，进一步提升航迹预测精度与收敛速率。

# 二、算法

蛇群算法（Snake Swarm Algorithm, SSA）是一种新型的元启发式算法，灵感来源于自然界中蛇群的觅食和生存行为，即蛇在自然环境中通过感知周围环境、追踪猎物以及与其他蛇协作来完成觅食和生存任务。[15]该算法通过模拟蛇群的这些行为模式和动态交互过程，提出了一种独特的模型，用于解决复杂的优化问题。下面开始详细介绍蛇群算法。

## 2.1 算法实现

### 2.1.1 初始化

与其他元启发式算法一样，蛇群算法首先会随机生成一个初始种群，然后开始算法的优化过程。蛇群的初始种群可以使用下式获得：

$$X_i = X_{min} + r \times (X_{max} - X_{min}) \qquad (1)$$

其中， $X_i$是第 i 个个体的位置，r 是介于 0 和 1 之间的随机数，$X_{max}$、$X_{min}$ 分别分别为问题位置的上限和下限。

### 2.1.2 划分雌雄两个子种群

在本文中，假设蛇群中雄性跟雌性的数量分别占 50%。再把蛇群划分为雄性跟雌性两个子种群，划分方式如下：

$$N_{male} = N/2 \qquad (2)$$

$$N_{female} = N_{all} - N_{male} \qquad (3)$$

其中，$N_{all}$是个体总数，$N_{male}$是雄性个体总数，$N_{female}$是雌性个体总数。

### 2.1.3 确定食物量与温度值并挑选最佳个体

分别评估两个子种群，挑选出最佳雄性个体跟最佳雌性个体，并确定食物位置。

温度值 Temp 可由以下公式确定：

$$Temp = \exp(\frac{-iter}{max_iter}) \qquad (4)$$

其中，iter 指当前迭代次数，max_iter 为最大迭代次数。

食物量 Q 可由以下公式确定：

$$Q = C_1 \times \exp(\frac{iter - max_iter}{max_iter}) \qquad (5)$$

其中，$C_1$为 0.5。

### 2.1.4 探索阶段

当蛇群缺乏食物并且 Q 小于阈值时，每个蛇个体会重新随机选择一个位置寻找食物，并更新自己的位置。 可通过如下公式模拟探索阶段：

$$X_{i,m}(t+1) = X_{rand,m}(t) \pm C_2 \times A_m \times ((X_{max} - X_{min}) \times rand + X_{min}) \qquad (6)$$

其中，$X_{i,m}$ 指第 i 个雄性个体的位置，$X_{rand,m}$ 指随机雄性个体的位置，rand 是一个 0-1 的随机值，$A_m$ 是雄性发现食物的能力，计算公式如下：

$$A_m = \exp(\frac{-f_{rand,m}}{f_{i,m}}) \tag{7}$$

其中，$f_{rand,m}$ 是 $X_{rand,m}$ 的适应度，$f_{i,m}$ 是雄性群体中第 i 个个体的适应度，C2 等于 0.05。

$$X_{i,f} = X_{rand,f}(t+1) \pm C_2 \times A_f \times ((X_{max} - X_{min}) \times rand + X_{min}) \tag{8}$$

其中，$X_{i,f}$ 指第 i 个雌性个体的位置，$X_{rand,f}$指随机雌性个体的位置，rand 是一个 0-1 的随机值，$A_f$是雌性发现食物的能力，计算公式如下：

$$A_f = \exp(\frac{-f_{rand,f}}{f_{i,f}}) \tag{9}$$

其中，$f_{rand,f}$是 $X_{rand,f}$的适应度，$f_{i,f}$ 是雌性群体中第 i 个个体的适应度。

### 2.1.5 开发阶段

当蛇群不缺乏食物并且 Q 大于阈值时：

（1）当环境温度大于阈值时，蛇群只会向食物移动：

$$X_{i,j}(t+1) = X_{food} \pm C_3 \times Temp \times rand \times (X_{food} - X_{i,j}(t)) \tag{10}$$

其中，$X_{i,j}$ 是个体（雄性或雌性）的位置，$X_{food}$是最佳个体的位置，c3 等于 2。

（2）当环境温度小于阈值时，蛇将会进入战斗模式或交配模式：

a. 战斗模式

$$X_{i,m}(t+1) = X_{i,m}(t) \pm C_3 \times FM \times rand \times (X_{best,f} - X_{i,m}(t)) \tag{11}$$

其中，$X_{i,m}$指第 i 个雄性个体的位置，$X_{best,f}$指雌性个体中最优秀个体的位置,$FM$是雄性个体的战斗力。

$$X_{i,f}(t+1) = X_{i,f}(t+1) \pm C_3 \times FF \times rand \times (X_{best,m} - X_{i,F}(t+1)) \tag{12}$$

其中，$X_{i,f}$指第 i 个雌性个体的位置，$X_{best,m}$指雄性个体中最优秀个体的位置，$FF$是雌性个体的战斗力。

FM 和 FF 可通过以下公式计算得出：

$$FM = \exp(\frac{-f_{best,f}}{f_i}) \tag{13}$$

$$FF = \exp(\frac{-f_{best,m}}{f_i}) \tag{14}$$

其中，$f_{best,f}$是雌性群体中最佳代理的适配度，$f_{best,m}$是雄性群体中最佳代理的适配度，$f_i$为个体适应度。

b. 交配模式

$$X_{i,m}(t+1) = X_{i,m}(t) \pm C_3 \times M_m \times rand \times (Q \times X_{i,f}(t) - X_{i,m}(t)) \tag{15}$$

$$X_{i,f}(t+1) = X_{i,f}(t) \pm C_3 \times M_f \times rand \times (Q \times X_{i,m}(t) - X_{i,f}(t)) \tag{16}$$

其中，$X_{i,f}$是第 i 个个体在雌性群体中的位置，$X_{i,m}$是第 i 个个体在雄性群体中的位置，$M_m$和$M_f$分别指雄性和雌性的交配能力。

计算公式如下：

$$M_m = \exp(\frac{-f_{i,f}}{f_{i,m}}) \tag{17}$$

$$M_f = \exp(\frac{-f_{i,m}}{f_{i,f}}) \tag{18}$$

如果孵化出蛋，则选择最差的雄性和雌性并替换它们

$$X_{worst,m} = M_{min} + rand \times (X_{max} - X_{min}) \tag{19}$$

$$X_{worst,f} = M_{min} + rand \times (X_{max} - X_{min}) \tag{20}$$

其中，$X_{worst,m}$是男性组中最差的个体，$X_{worst,f}$是女性组中最差的个体。

## 2.2 算法改进

### 2.2.1 佳点集初始化

佳点集是一种具有优良分布特性的离散点集，其基于数论中的均匀分布理论构建。在群智能优化算法初始化过程中，相较于传统随机初始化方法，采用佳点集初始化策略能够使种群个体在搜索空间中呈现更均匀、更具代表性的分布。[16]该策略通过选取合适的生成点，依据特定的数学规则生成点集，有效避免了种群聚集现象，增强了算法初始阶段的全局探索能力，为后续迭代寻优过程奠定更优的初始种群基础，从而提升算法的收敛速度与寻优精度。本研究引入了佳点集初始化以提升初始种群的质量，并使得研究结果更具普适性，具体方法如下：

假设在单位立方体中存在点集：

$$P_n(k) = \left\{ \left( \left\{ r_1^{(n)} \cdot k \right\}, \left\{ r_2^{(n)} \cdot k \right\}, \left\{ r_3^{(n)} \cdot k \right\}, \ldots\ldots, \left\{ r_s^{(n)} \cdot k \right\} \right), 1 \le k \le n \right\} \tag{21}$$

其偏差满足：

$$\varphi_n = C(r, \varepsilon) n^{-1+\varepsilon} \tag{22}$$

其中，$C(r, \varepsilon)$是常数，只与 $r, \varepsilon$ $(\varepsilon > 0)$有关。那么$P_n(k)$称为佳点集，r 也称为佳点集。佳点集 r 的值为：

$$r_i = 2\cos\frac{2\pi i}{p}, 1 \le i \le s \tag{23}$$

其中，p 是满足 $\frac{(p-3)}{2} > s$ 的最小素数。

基于佳点集公式，可以将蛇群的初始化改进为以下策略：

$$X_i(k) = (X_{upper} - X_{low})\{P_n(k)\} + X_{low} \tag{24}$$

其中，$X_{upper}$为空间范围的上界，$X_{low}$为空间范围的下界

### 2.2.2 阈值参数自适应化

鉴于更好模拟蛇群生活习性与提高现实问题优化效果的需要，本文在 C1、C3、食物阈值、温度阈值四个参数上进行了基于周期性变化的自适应改进。[17]参数的自适应化既能满足实际工程问题的需要，周期化的变化规律又能避免波动过大化，符合自然界潜移默化的演变规律。[18][19]

(a)$C_1$的自适应化设计

$C_1$参数用于控制蛇群个体对全局最优解的吸引力，本文通过余弦函数的周期性变化，使$C_1$在 [0, 1] 范围内波动，动态调整蛇群个体对全局最优解的追随力度，自适应公式如下：

$$C_1 = 0.5 \times (1 + \cos\frac{2\pi \cdot iter}{T}) \quad (25)$$

其中，T 为周期参数，公式如下：

$$oT = \frac{max_iter}{2} \quad (26)$$

(b)$C_3$的自适应化设计

$C_3$参数用于控制蛇群个体对其他蛇（局部最优解）的排斥力,本文通过正弦函数的周期性变化，使$C_3$在 [2, 4] 范围内波动，动态调整蛇群个体对局部最优解的排斥力度，自适应公式如下：

$$C_3 = 2 \times (1 + \sin\frac{2\pi \cdot iter}{T}) \quad (27)$$

(c)food_threshold 的自适应化设计

food_threshold 参数用于控制蛇群对食物（全局最优解）的追逐行为，本文通过正弦函数的周期性变化，使 food_threshold 在 [0.25, 0.75] 范围内波动，动态调整蛇群对全局最优解的依赖程度，自适应公式如下：

$$food_threshold = 0.5 + 0.25 \cdot \sin\frac{2\pi \cdot iter}{T} \quad (28)$$

(d)temp_threshold 的自适应化设计

temp_threshold 参数决定了蛇群下一步的行为模式，用于控制蛇群是靠近食物还是战斗繁衍，本文通过余弦函数的周期性变化，使 temp_threshold 在 [0.3, 0.7]] 范围内波动，动态调整蛇群的多样性和收敛性，自适应公式如下：

$$temp_threshold = 0.5 + 0.2 \cdot \cos\frac{2\pi \cdot iter}{T} \quad (29)$$

## 2.2.3 种群双重变异策略

本文设计了一种基于主辅变异方案结合的双重变异策略。第一重变异为主要变异方案，旨在根据种群迭代的阶段来平衡全局搜索能力跟局部搜索能力；第二重变异为辅助变异方案，旨在通过不同的个体变异方式来增加种群多样性，避免算法陷入局部最优。

（1）主要变异方案

选用高斯变异和柯西变异作为主要变异策略，在算法早期应用柯西变异，以增强全局搜索能力，在算法后期应用高斯变异，以增强局部搜索能力。[20][21]

a. 柯西变异

针对蛇群算法在探索阶段易陷入局部最优值的缺陷，引入柯西噪声对蛇个体的位置$X$进行扰动，以更新蛇的最新位置。柯西分布的重尾特性使其更有可能产生较大的变异步长，从而帮助算法快速穿越搜索空间，避免过早收敛到局部最优解。[22]基于柯西变异的蛇群位置更新公式如下：

$$X' = X + Cauchy(0, \gamma) \quad (30)$$

其中，$X$是蛇的当前位置，$X'$是变异后的新的位置向量，$\gamma$为常数 0.05，$Cauchy(0,\gamma)$表示位置参数为 0、尺度参数为$\gamma$的柯西分布。

b. 高斯变异

针对蛇群算法在开发阶段对于最佳区域开发不够精细的问题，引入高斯噪声对蛇个体的

最佳位置进行微细调整，以获得蛇的最优位置。高斯分布较小的变异步长有助于算法在最有希望的区域进行精细调整，从而提高解决方案的质量并最终收敛到全局最优解。基于高斯变异的蛇群位置更新公式如下：

$$X^{'} = X + N(0, \sigma^2) \tag{31}$$

其中，$\sigma$为常数 0.1，$N(0,\sigma^2)$表示均值为 0、方差为$\sigma^2$的高斯分布。

（2）辅助变异方案

选用头部混沌变异、身体融合变异、尾部拼接变异三种作为辅助变异策略，分别对蛇个体的头、身、尾三个部位进行变异操作，以最大化地丰富种群的多样性，从而找到最佳个体。

c. 头部混沌变异

在本算法中，通过 Logistic 混沌映射引入非线性动态行为，实现对蛇头部位置的变异更新。该策略通过引入混沌行为，增强了算法的全局搜索能力，使蛇能够跳出局部最优解，探索更广泛的搜索空间。具体而言，对于蛇的当前位置向量 x，我们通过以下公式进行变异：

$$X^{'} = X + \alpha \cdot (X \cdot (1 - X)) \tag{32}$$

其中，α是混沌因子，控制混沌映射的强度。

d. 身体融合变异

身体融合变异通过将两条蛇的位置向量在中点处分割并拼接，生成新的位置向量。该策略通过融合两条蛇的特征，增强了算法的多样性和探索能力，使蛇群能够更好地适应复杂的搜索空间。具体而言，对于两条蛇的位置向量$X_1$和$X_2$，我们通过以下公式进行变异：

$$X^{'} = \frac{1}{2}(X_1 + X_2) \tag{33}$$

其中，$X_1$为需要变异的蛇的位置向量，$X_2$为随机选择的另一条蛇的位置向量。

e. 尾部拼接变异

尾部拼接变异通过将一条蛇的尾部与另一条蛇的头部拼接，生成新的位置向量。该策略通过重新组合蛇的位置向量，增强了算法的多样性和探索能力，使蛇群能够更好地适应复杂的搜索空间。具体而言，对于两条蛇的位置向量$X_1$和$X_2$，我们通过以下公式进行变异：

$$X^{'} = \begin{bmatrix} X_1[m:] \\ X_2[:m] \end{bmatrix} \tag{34}$$

其中，m 是蛇个体尾部的起始点。

### 2.2.4 飞行函数引入

在本文中，采用了一种改进的自适应策略，将增强的 Levy 飞行函数（adaptive_Levy flight）与基于随机游走的飞行函数（random_walk_flight）相结合。[23][24]具体而言，在算法的早期阶段，我们优先采用 Levy 飞行函数，以增强算法的全局搜索能力，从而更有效地探索解空间的广泛区域。随着算法的迭代进行，进入后期阶段后，我们则切换至随机游走函数，以强化算法的局部搜索能力，实现对潜在最优解的精细搜索。通过这种自适应的切换机制，算法能够在全局探索与局部开发之间实现有效的平衡，从而提高求

解复杂优化问题的效率和精度。[25]

（1）adaptive_Levy flight

自适应 Levy 飞行函数在算法的早期阶段促进全局探索，在后期阶段平滑过渡到局部探索。步长根据迭代次数自适应调整：

$$\text{adaptive_Levy_flight}(d, iter, max_iter) = \text{Levy 步长} \times \left(1 - \frac{iter}{max_iter}\right)^{\beta} \quad (35)$$

其中，d 是维度，β是控制步长减小速度的参数，Levy 步长公式如下：

$$\text{Levy 步长} = \frac{u}{|v|^{\frac{1}{\beta}}} \quad (36)$$

（2）random_walk flight

随机游走飞行函数在算法的后期阶段促进局部探索,公式如下：

$$\text{random_walk_flight}(d) = \mu(-\sigma, \sigma) \quad (37)$$

其中，$\mu(-\sigma,\sigma)$表示在 $[-\sigma,\sigma]$ 范围内均匀分布的随机数。

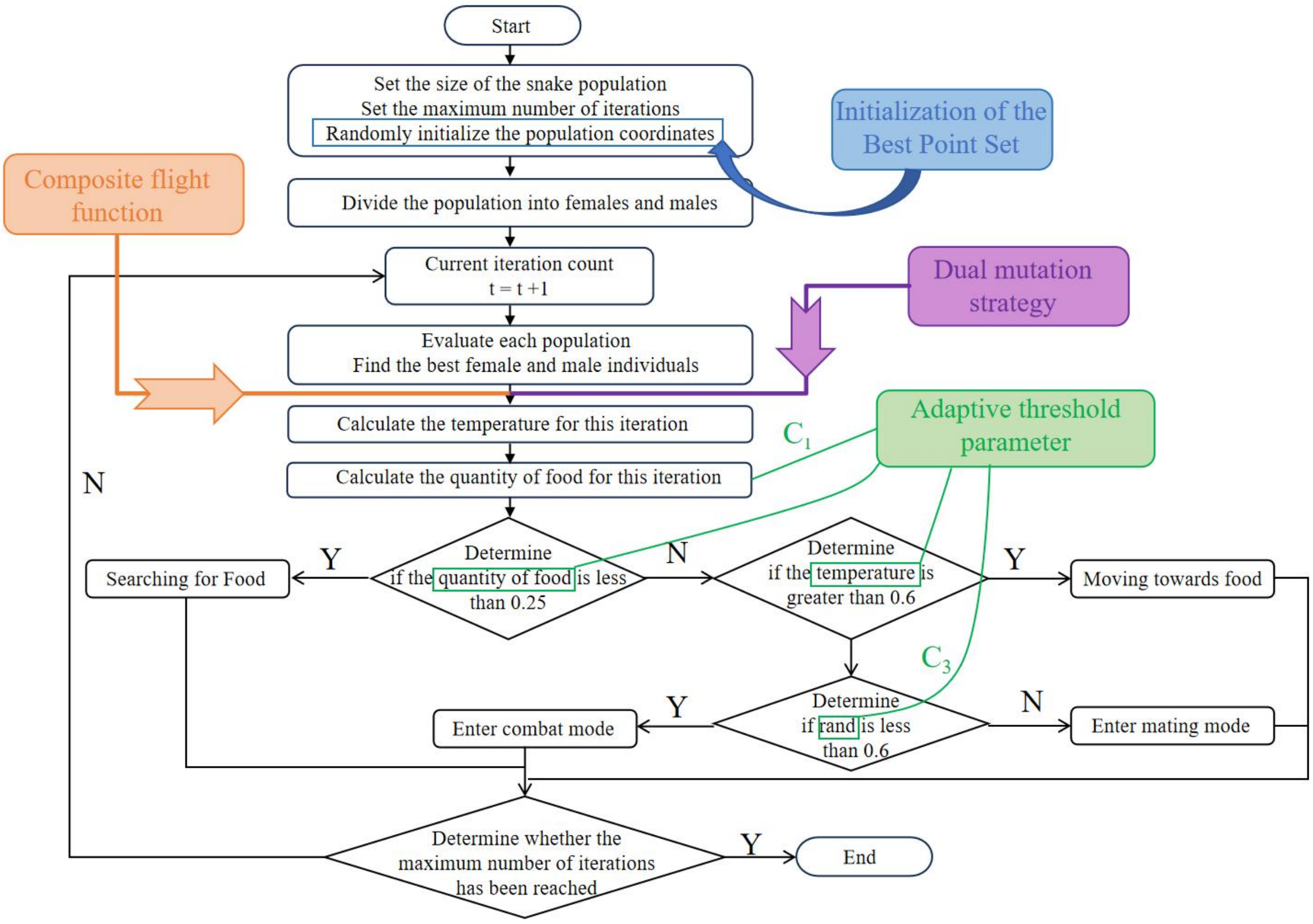

Figure1 Flowchart of SO algorithm based on multi-strategy improvement

## 2.3 算法性能测试与分析

### 2.3.1 性能测试设置

为了全面客观的评估基于多策略改进后的 SO 算法的性能，本小节将 SO 与包括 WOA、GWO、BSO 等在内的八种优秀启元式优化算法在 cec2022 测试集中的 12 种不同种类的基准测试函数上进行比较，种群数和迭代次数分别统一设置为 30 和 500。

硬件环境为 Intel® Xeon® CPU E5-2678 v3 @ 2.50GHz × 48 核，62.8GB 内存及 NVIDIA GeForce RTX 4090 Ti 的 64 位计算机。

软件平台为 Windows11 64bit 操作系统，Python 3.6 及 TensorFlow 2.10.0 深度学习框架。在训练深度学习模型时使用 TensorFlow 的 GPU 版本，利用 GPU 进行硬件加速。

挑选的 12 个测试函数如表 1 所示，涵盖了多种不同的维度、定义域和最优解，其中，F1 为单峰测试函数；F2 至 F5 为基础测试函数；F6 至 F8 为混合测试函数；F9 至 F12 为组合测试函数，测试维度统一设置为 20。[26]

Table1 Test Functions

| Function Category | Serial Number | Function | Dim | optimal value |
|---|---|---|---|---|
| Unimodal Function | F1 | Shifted and full Rotated Zakharov Function | 20 | 300 |
| Basic Function | F2 | Shifted and full Rotated Rosenbrock's Function | 20 | 400 |
| | F3 | Shifted and full Rotated Expanded Schaffer's f6 Function | 20 | 600 |
| | F4 | Shifted and full Rotated Levy Function | 20 | 900 |
| | F5 | Hybrid Function 2 (N=6) | 20 | 2000 |
| Composition Functions | F6 | Composition Function 1 (N=3) | 20 | 2200 |
| | F7 | Composition Function 1 (N=5) | 20 | 2300 |
| | F8 | Composition Function 2 (N=4) | 20 | 2400 |
| | F9 | Composition Function 3 (N=5) | 20 | 2600 |
| | F10 | Composition Function 4 (N=6) | 20 | 2700 |

Search range: $[-100,100]^D$

### 2.3.2 测试结果比较

SO 算法与八种对比算法的测试实验结果如表 2 所示：

在单峰函数的实证测试中，SO 算法在平均值上达到了最优解，与 BSO、PSO 算法并列第一，展现出了强大的局部搜索能力；

在基础函数的实证测试中，SO 算法在 F2、F3、F5 上取得了第一，仅在 F4 上以微小的差距略逊色于 BSO 算法；

在混合函数的实证测试中，SO 算法在 F7、F8 上取得了第一，而在 F6 上因测试得出的最优值不够理想，表现差于 BSO、GWO 算法位居第三；

在组合函数的实证测试中，SO 算法以较大的领先优势断层式取得了第一，展现出了及其优越的全局搜索能力与跳出局部最优解的能力。

在 12 项不同维度的测试函数的测试下，SO 算法一共在十项测试上获得了第一，展现了经过多策略改进的 SO 算法强大的高效性与稳定性，这种性能优势可归因于多样化的改进策略，这些机制共同提升了算法的全局搜索效率，并有效的减少了陷入局部最优的风险。

Table2 Comparison of PLORCS and classical optimization algorithms for finding the best

| 测试函数 | 统计值 | WOA | GWO | BSO | PSO | DE | GA | BBO | GCO | SO |
|---|---|---|---|---|---|---|---|---|---|---|
| F1 | Optimal Value | 4.02E+02 | 3.11E+02 | 3.00E+02 | 3.00E+02 | 5.88E+03 | 4.85E+03 | 1.04E+03 | 2.22E+03 | 3.00E+02 |
| | Worst Value | 5.24E+03 | 3.84E+02 | 3.00E+02 | 3.00E+02 | 1.610E+04 | 6.81E+03 | 6.07E+03 | 2.75E+03 | 3.02E+02 |
| | Average Value | 2.68E+03 | 3.46E+02 | 3.00E+02 | 3.00E+02 | 1.11E+04 | 5.85E+03 | 2.79E+03 | 2.52E+03 | 3.00E+02 |
| | Standard Deviation | 1.77E+03 | 2.49E+01 | 1.11E-01 | 5.08E-14 | 4.26E+03 | 6.26E+02 | 1.73E+03 | 1.90E+02 | 8.32E-01 |
| | Ranking | 6 | 4 | 1 | 1 | 9 | 8 | 7 | 5 | 1 |
| F2 | Optimal Value | 5.34E+02 | 4.49E+02 | 4.63E+02 | 4.30E+02 | 1.12E+03 | 1.66E+03 | 4.54E+02 | 7.42E+02 | 4.49E+02 |
| | Worst Value | 7.24E+02 | 5.19E+02 | 4.83E+02 | 4.73E+02 | 1.67E+03 | 2.01E+03 | 4.97E+02 | 1.17E+03 | 4.59E+02 |
| | Average Value | 6.55E+02 | 4.87E+02 | 4.71E+02 | 4.55E+02 | 1.33E+03 | 1.80E+03 | 4.72E+02 | 9.38E+02 | 4.54E+02 |
| | Standard Deviation | 6.47E+01 | 2.32E+01 | 7.78E+00 | 1.81E+01 | 2.02E+02 | 1.47E+02 | 1.8E+01 | 1.56E+02 | 3.18E+00 |
| | Ranking | 6 | 5 | 3 | 2 | 8 | 9 | 4 | 7 | 1 |
| F3 | Optimal Value | 6.00E+02 | 6.00E+02 | 6.00E+02 | 6.00E+02 | 6.00E+02 | 6.00E+02 | 6.00E+02 | 6.00E+02 | 6.00E+02 |
| | Worst Value | 6.00E+02 | 6.00E+02 | 6.00E+02 | 6.00E+02 | 6.01E+02 | 6.01E+02 | 6.00E+02 | 6.00E+02 | 6.00E+02 |
| | Average Value | 6.00E+02 | 6.00E+02 | 6.00E+02 | 6.00E+02 | 6.01E+02 | 6.01E+02 | 6.00E+02 | 6.00E+02 | 6.00E+02 |
| | Standard Deviation | 2.17E-01 | 1.29E-02 | 7.70E-05 | 3.19E-05 | 1.14E-01 | 4.84E-02 | 4.08E-04 | 4.00E-02 | 7.03E-04 |
| | Ranking | 1 | 1 | 1 | 1 | 8 | 8 | 1 | 1 | 1 |
| F4 | Optimal Value | 9.06E+02 | 9.00E+02 | 9.02E+02 | 9.01E+02 | 9.04E+02 | 9.06E+02 | 9.01E+02 | 9.13E+02 | 9.00E+02 |
| | Worst Value | 9.10E+02 | 9.06E+02 | 9.05E+02 | 9.07E+02 | 9.07E+02 | 9.13E+02 | 9.05E+02 | 9.22E+02 | 9.00E+02 |
| | Average Value | 9.07E+02 | 9.01E+02 | 9.03E+02 | 9.04E+02 | 9.05E+02 | 9.11E+02 | 9.03E+02 | 9.17E+02 | 9.00E+02 |
| | Standard Deviation | 1.43E+00 | 2.20E+00 | 1.09E+00 | 2.14E+00 | 1.16E+00 | 2.64E+00 | 1.45E+00 | 3.13E+00 | 8.43E-02 |
| | Ranking | 7 | 2 | 3 | 5 | 6 | 8 | 3 | 9 | 1 |
| F5 | Optimal Value | 2.29E+03 | 2.06E+03 | 2.11E+03 | 2.12E+03 | 2.13E+03 | 3.10E+03 | 2.04E+03 | 2.48E+03 | 2.00E+03 |
| | Worst Value | 4.11E+03 | 2.12E+03 | 2.69E+03 | 2.56E+03 | 3.60E+03 | 3.94E+03 | 2.16E+03 | 3.92E+03 | 2.05E+03 |
| | Average Value | 3.31E+03 | 2.09E+03 | 2.48E+03 | 2.27E+03 | 2.96E+03 | 3.53E+03 | 2.07E+03 | 3.15E+03 | 2.02E+03 |
| | Standard Deviation | 5.83E+02 | 2.29E+01 | 2.16E+02 | 1.51E+02 | 5.05E+02 | 2.72E+02 | 4.46E+01 | 4.70E+02 | 1.28E+01 |
| | Ranking | 9 | 2 | 5 | 4 | 6 | 8 | 3 | 7 | 1 |
| F6 | Optimal Value | 2.42E+03 | 2.32E+03 | 2.41E+03 | 2.32E+03 | 2.38E+03 | 2.46E+03 | 2.32E+03 | 2.37E+03 | 2.32E+03 |
| | Worst Value | 2.68E+03 | 2.33E+03 | 2.93E+03 | 2.37E+03 | 2.55E+03 | 2.52E+03 | 2.33E+03 | 2.40E+03 | 2.33E+03 |
| | Average Value | 2.51E+03 | 2.33E+03 | 2.67E+03 | 2.34E+03 | 2.44E+03 | 2.49E+03 | 2.32E+03 | 2.38E+03 | 2.32E+03 |
| | Standard Deviation | 9.34E+01 | 4.71E+00 | 1.86E+02 | 1.37E+01 | 6.58E+01 | 2.48E+01 | 1.40E+00 | 1.33E+01 | 1.34E+00 |
| | Ranking | 8 | 3 | 9 | 4 | 6 | 7 | 1 | 5 | 1 |
| F7 | Optimal Value | 2.76E+03 | 2.66E+03 | 2.73E+03 | 2.68E+03 | 3.09E+03 | 3.24E+03 | 2.69E+03 | 2.66E+03 | 2.64E+03 |
| | Worst Value | 3.88E+03 | 2.76E+03 | 2.80E+03 | 2.78E+03 | 4.07E+03 | 3.54E+03 | 2.75E+03 | 2.70E+03 | 2.65E+03 |
| | Average Value | 3.10E+03 | 2.68E+03 | 2.77E+03 | 2.73E+03 | 3.47E+03 | 3.39E+03 | 2.72E+03 | 2.69E+03 | 2.64E+03 |
| | Standard Deviation | 4.00E+02 | 3.74E+01 | 3.19E+01 | 3.87E+01 | 3.47E+02 | 1.17E+02 | 2.10E+01 | 1.40E+01 | 5.09E+00 |
| | Ranking | 7 | 2 | 6 | 5 | 9 | 8 | 4 | 3 | 1 |
| F8 | Optimal Value | 2.93E+03 | 2.81E+03 | 4.61E+03 | 2.81E+03 | 3.08E+03 | 2.96E+03 | 2.81E+03 | 3.04E+03 | 2.68E+03 |
| | Worst Value | 6.90E+03 | 6.29E+03 | 6.37E+03 | 6.78E+03 | 5.52E+03 | 3.18E+03 | 3.63E+03 | 7.10E+03 | 3.19E+03 |
| | Average Value | 5.07E+03 | 4.71E+03 | 5.36E+03 | 4.67E+03 | 4.28E+03 | 3.07E+03 | 3.08E+03 | 4.62E+03 | 2.88E+03 |
| | Standard Deviation | 1.36E+03 | 1.52E+03 | 6.80E+02 | 1.31E+03 | 9.73E+02 | 7.65E+01 | 3.37E+02 | 1.78E+03 | 6.23E+01 |
| | Ranking | 8 | 7 | 9 | 6 | 4 | 2 | 3 | 5 | 1 |
| F9 | Optimal Value | 2.71E+03 | 2.60E+03 | 2.60E+03 | 2.60E+03 | 3.75E+03 | 3.07E+03 | 2.61E+03 | 4.17E+03 | 2.60E+03 |
| | Worst Value | 5.17E+03 | 2.63E+03 | 2.63E+03 | 2.62E+03 | 1.06E+04 | 3.92E+03 | 2.61E+03 | 4.27E+03 | 2.61E+03 |
| | Average Value | 3.67E+03 | 2.62E+03 | 2.61E+03 | 2.61E+03 | 5.63E+03 | 3.51E+03 | 2.61E+03 | 4.22E+03 | 2.61E+03 |
| | Standard Deviation | 9.61E+02 | 7.53E+00 | 1.14E+01 | 8.19E+00 | 2.61E+03 | 2.99E+02 | 2.32E+00 | 4.31E+01 | 3.86E+00 |
| | Ranking | 7 | 5 | 1 | 1 | 9 | 6 | 1 | 8 | 1 |
| F10 | Optimal Value | 3.08E+03 | 2.96E+03 | 3.63E+03 | 2.98E+03 | 3.11E+03 | 3.19E+03 | 2.95E+03 | 2.95E+03 | 2.94E+03 |
| | Worst Value | 3.78E+03 | 3.02E+03 | 3.99E+03 | 3.11E+03 | 3.32E+03 | 3.28E+03 | 3.06E+03 | 3.00E+03 | 2.95E+03 |
| | Average Value | 3.39E+03 | 2.98E+03 | 3.79E+03 | 3.03E+03 | 3.20E+03 | 3.24E+03 | 2.99E+03 | 2.98E+03 | 2.94E+03 |
| | Standard Deviation | 2.50E+02 | 2.01E+01 | 1.36E+02 | 4.43E+01 | 8.07E+01 | 4.08E+01 | 3.69E+01 | 2.10E+01 | 4.61E+00 |
| | Ranking | 8 | 2 | 9 | 5 | 6 | 7 | 4 | 2 | 1 |

### 2.3.3 收敛曲线及箱线图分析

如图所展示的，SO 算法在 12 项测试函数的测试下表现出了显著的优越性，其收敛曲线不仅光滑平整，而且在大多数情况下最终的收敛值要低于其他算法曲线，这表明 SO 算法在收敛性能上拥有着相对于其他算法显著的优越性。此外，SO 算法的收敛曲线往往在较短的时间内便能收敛到最优值，意味着其拥有快于其他算法的收敛速度。

如图所展示的，SO 算法在测试中有着最低的平均值，说明其具有良好的寻优效果。此外，SO 算法的二十次测试的结果分布总体上最为均衡，标准差是对比算法中最小的，说明其具有优良的稳定性。

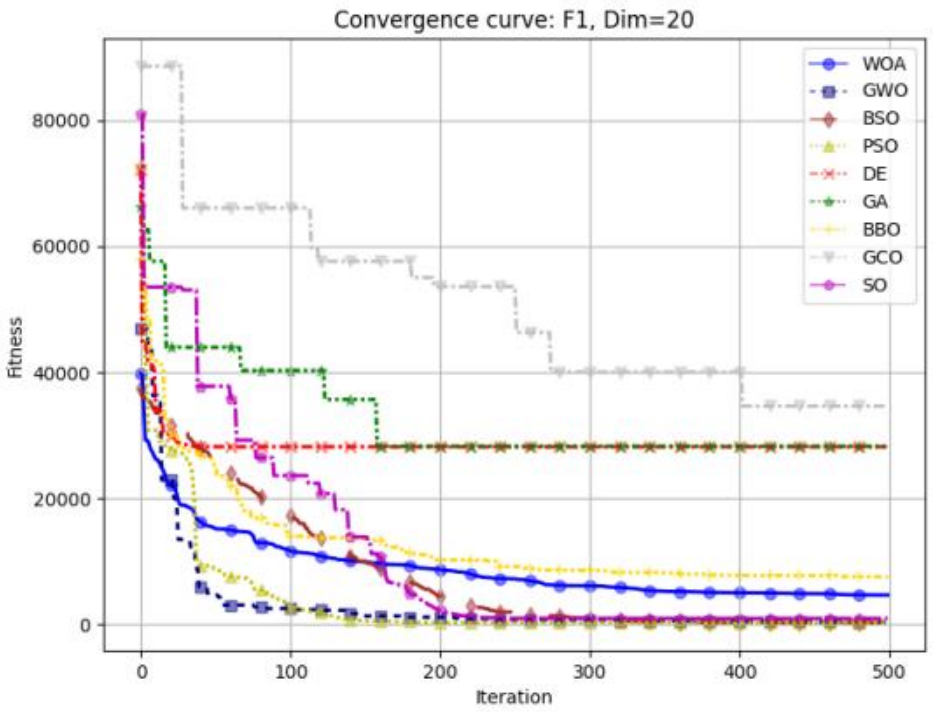

(a)

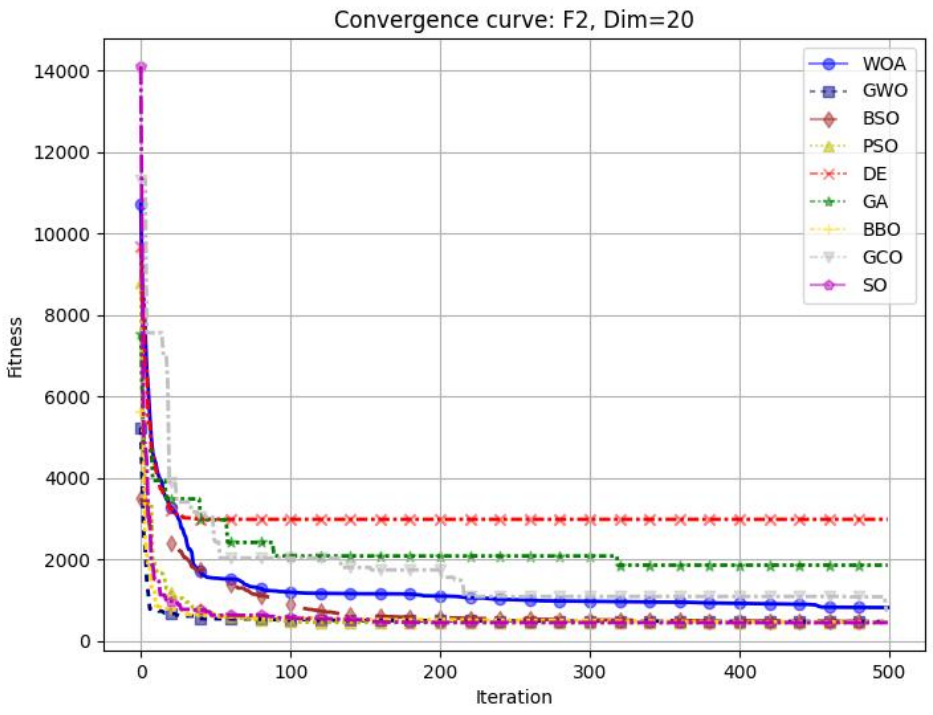

(b)

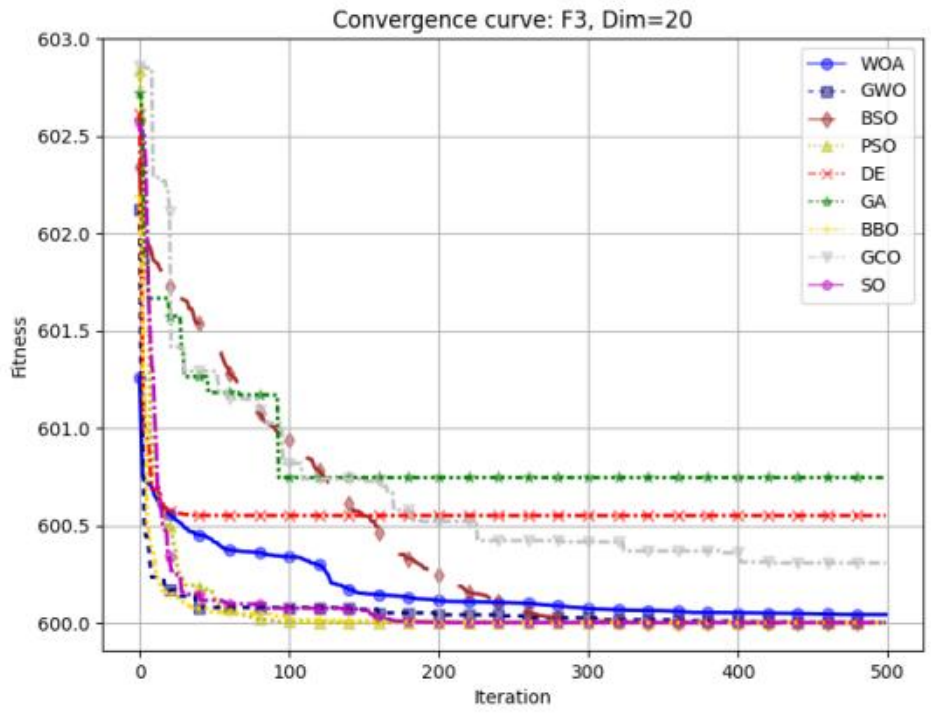

(c)

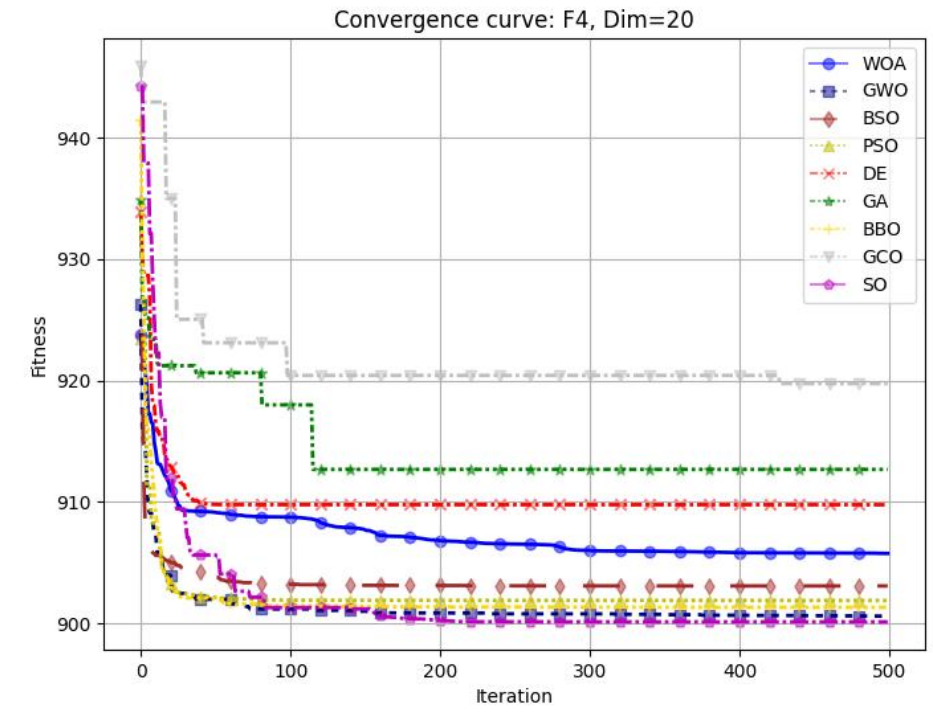

(d)

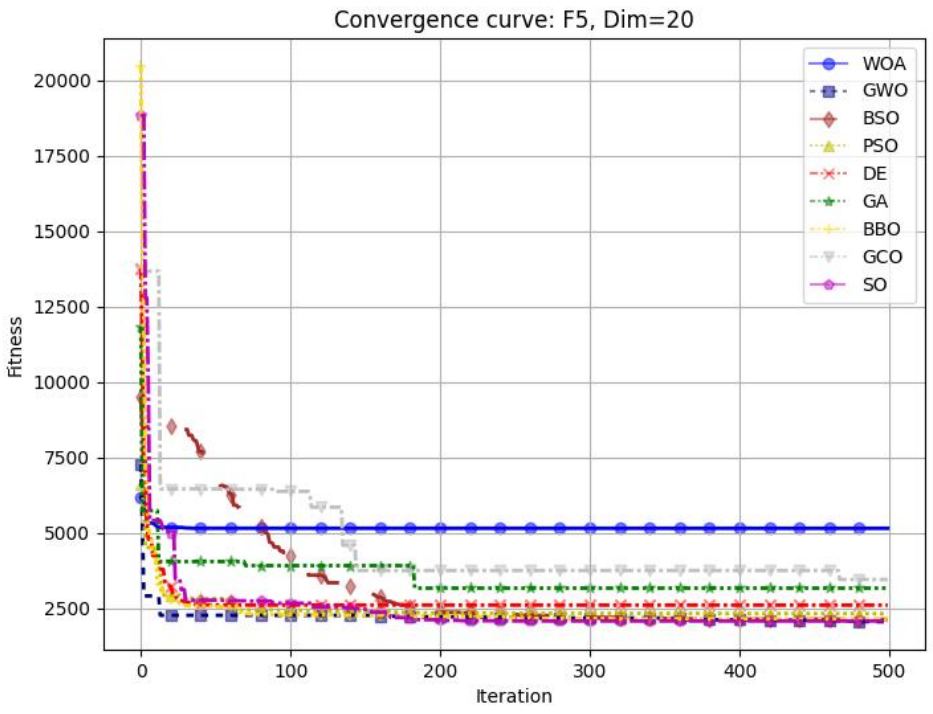

(e)

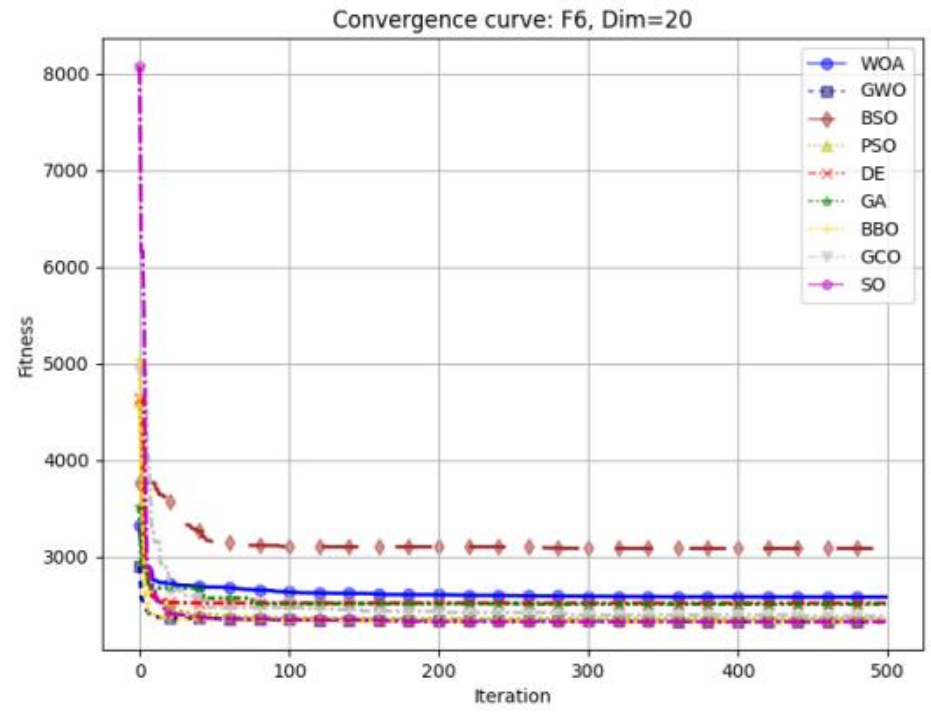

(f)

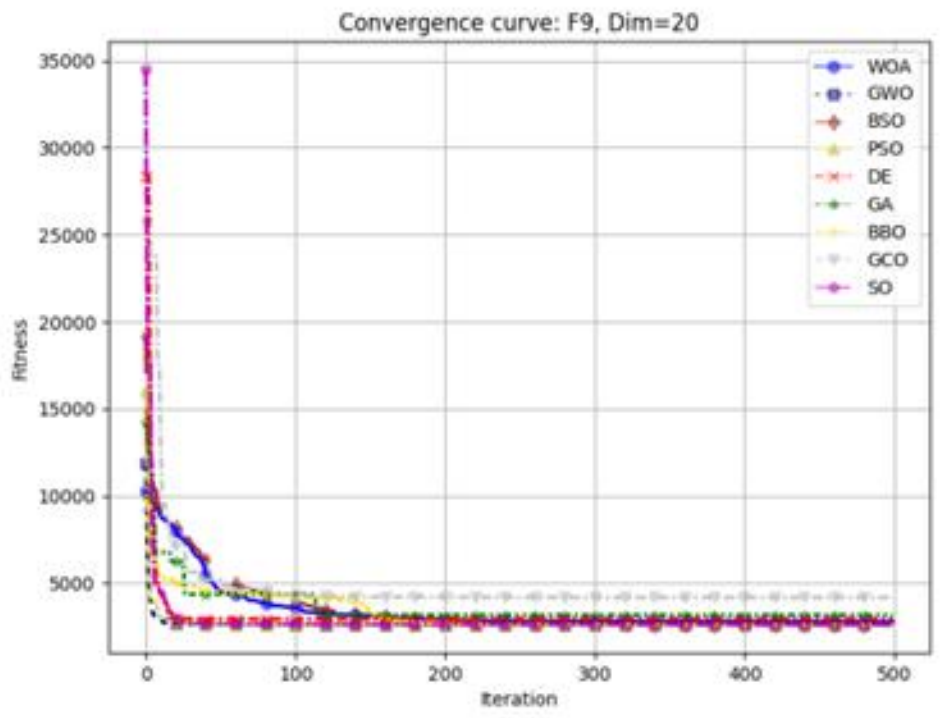

(g)

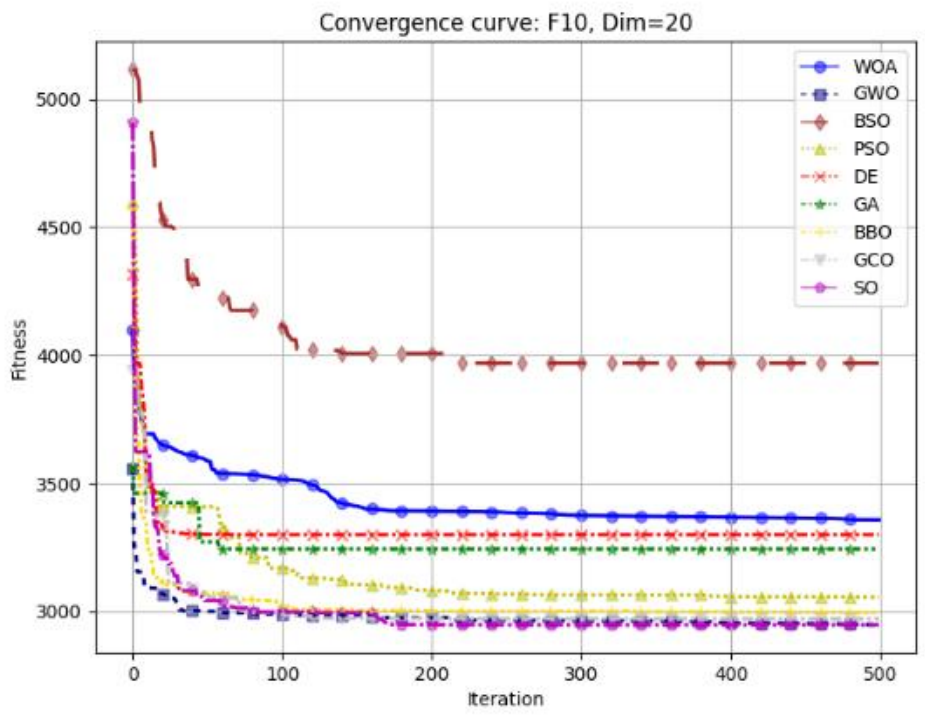

(h)

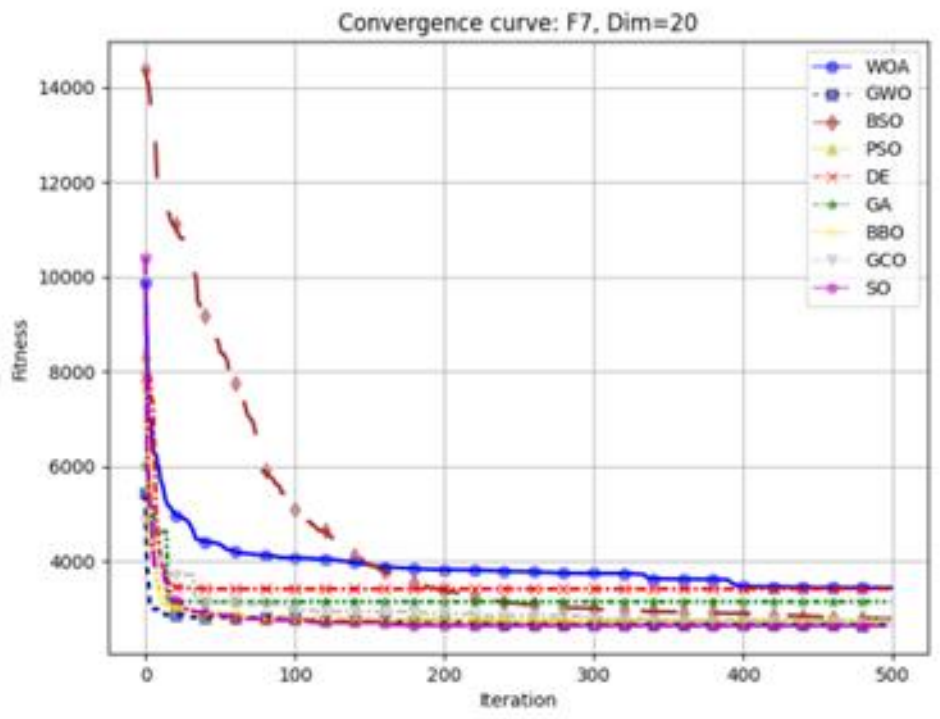

(i)

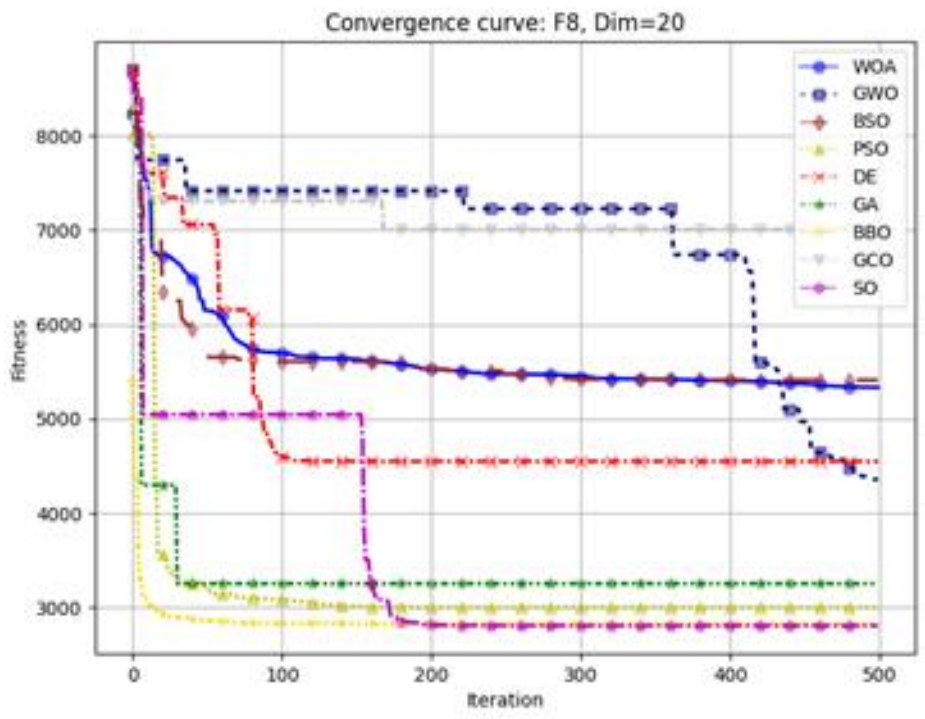

(j)

Figure2 Test function convergence curve

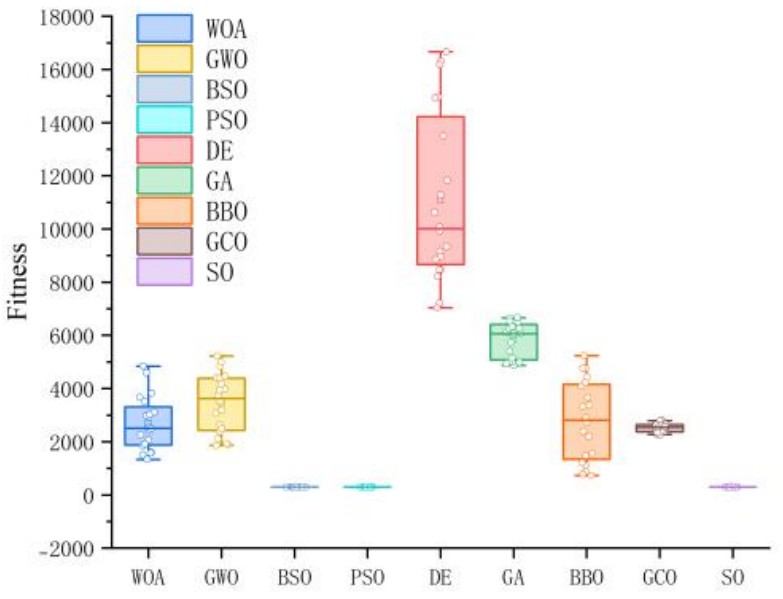

(a)

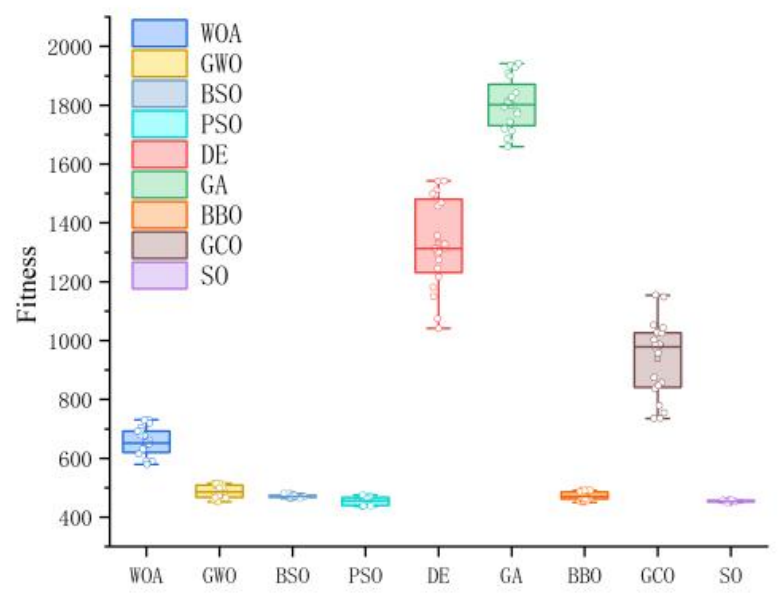

(b)

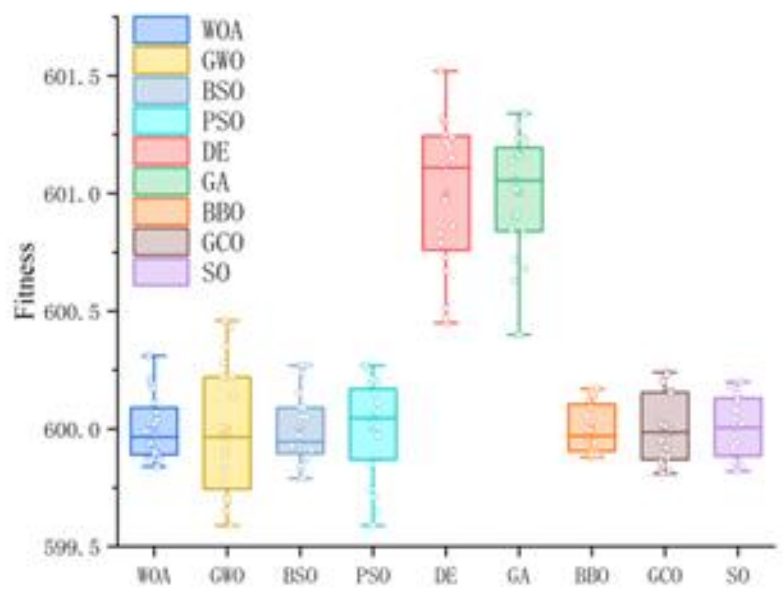

(c)

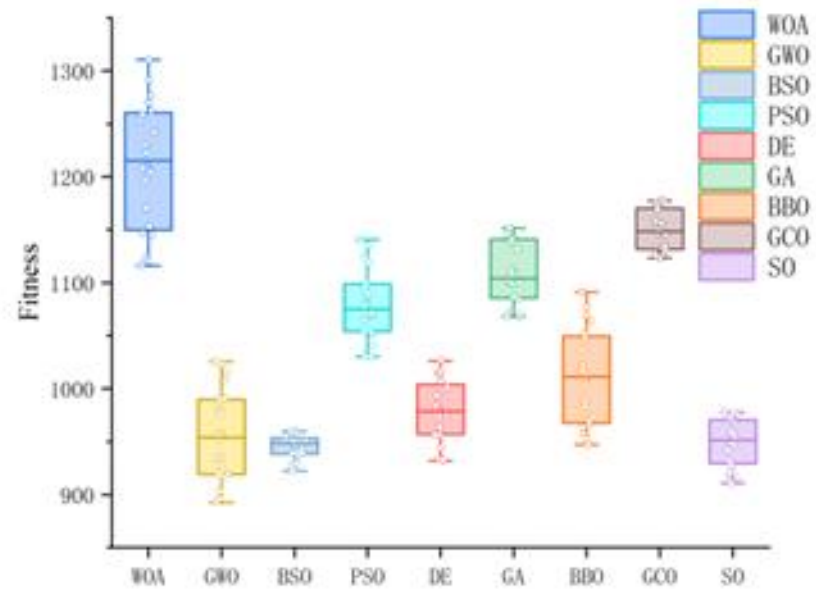

(d)

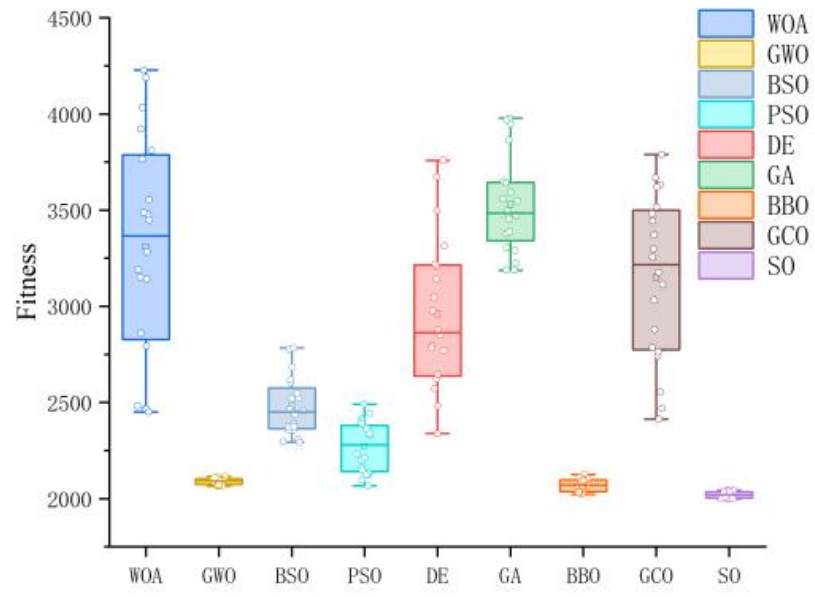

(e)

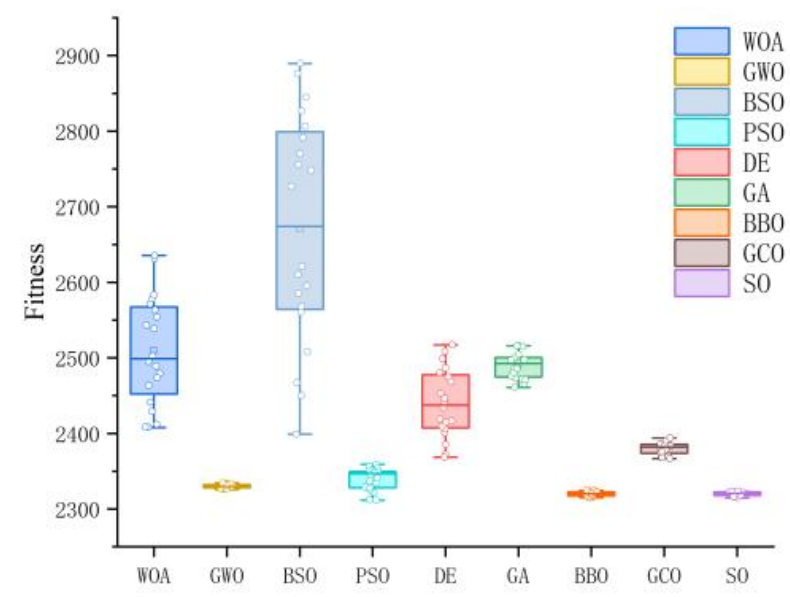

(f)

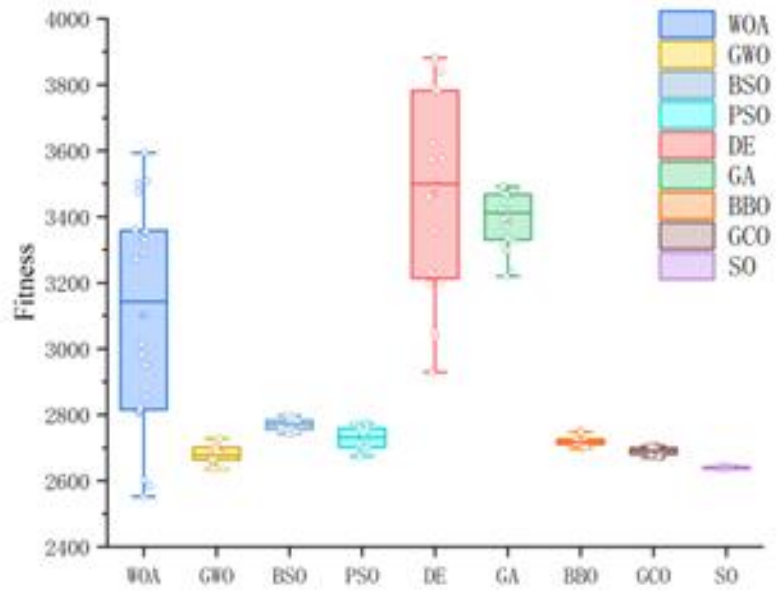

(g)

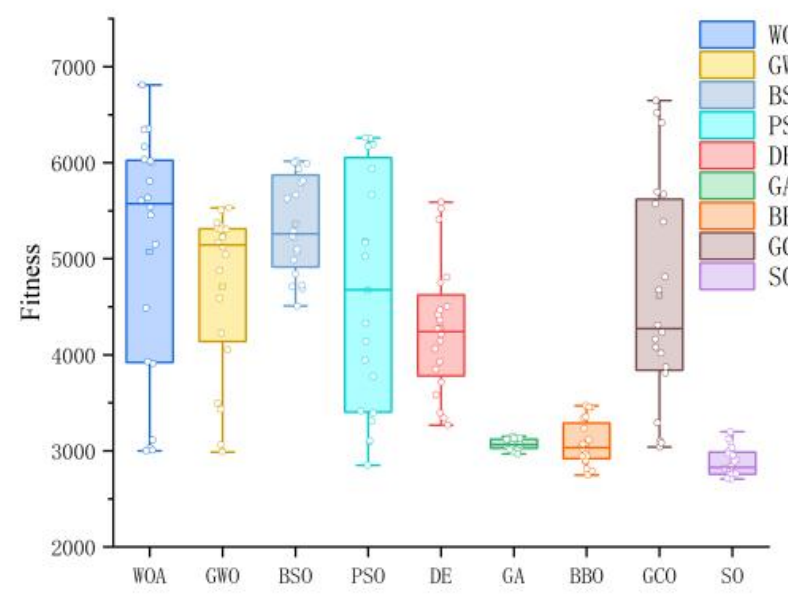

(h)

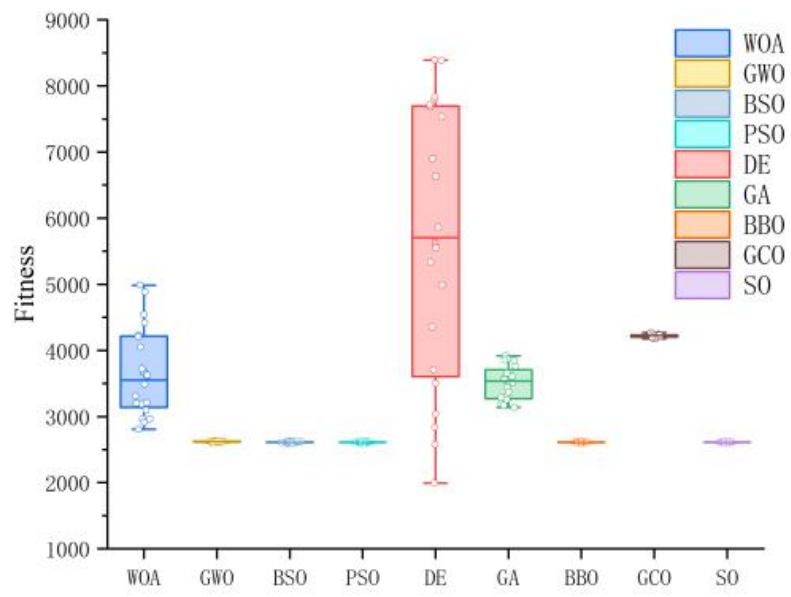

(i)

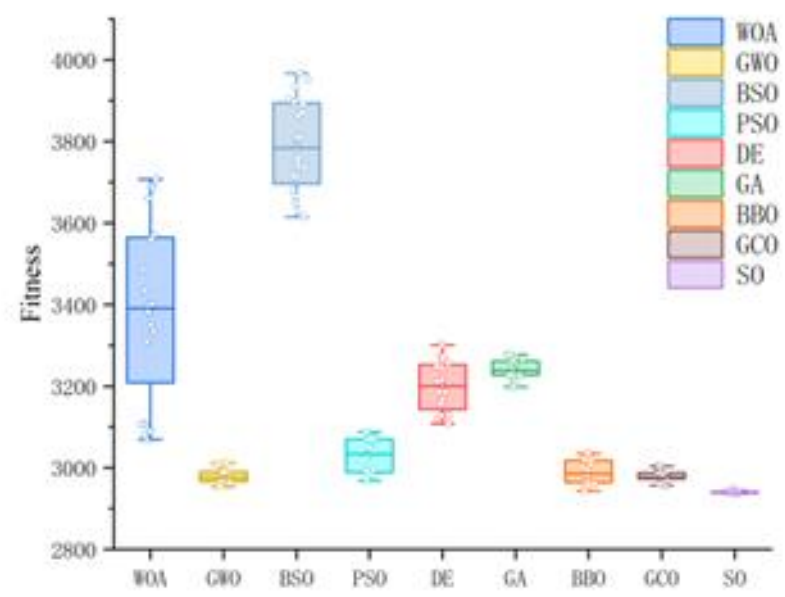

(j)

Figure3 Test function box diagram

## 2.4 Wilcoxon 秩和检验分析

如表 3 所示，本研究引入了 Wilcoxon 秩和检验，利用 p 值来验证改进后的 SO 与其他五种算法是否具有显著差异性。[27]若 $p < 0.05$ 时，两种算法之间具有显著差异性。“+/=/–”分别表示改进后的 SO 在性能上“优于/相当/逊于”其他算法。

由表 3 可知，改进后的 SO 在 F1-F10 的 p 值多数小于 0.05 且为“+”，说明其在测试函数上的表现比其他 8 种算法性能更优，其中：PSO 与 SO 在 F1-F3 的全局寻优上性能相当，皆能取得最优值；BSO 与 BBO 分别在 F1 与 F6 上的测试性能与 SO 相当。

Table3　p-value of Wilcoxon rank sum test

| Test | WOA | GWO | BSO | PSO | DE | GA | BBO | GCO |
|---|---|---|---|---|---|---|---|---|
| F1 | 1.907E-06 | 1.907E-06 | 8.124E-01 | 8.124E-01 | 1.907E-06 | 1.907E-06 | 1.907E-06 | 1.907E-06 |
| F2 | 1.907E-06 | 1.907E-06 | 1.907E-06 | 7.012E-01 | 1.907E-06 | 1.907E-06 | 3.815E-06 | 1.907E-06 |
| F3 | 7.557E-10 | 7.557E-10 | 7.557E-10 | 3.771E-01 | 7.557E-10 | 7.557E-10 | 9.634E-10 | 7.557E-10 |
| F4 | 7.523E-10 | 2.425E-08 | 7.523E-10 | 7.555E-10 | 7.550E-10 | 7.540E-10 | 7.555E-10 | 7.550E-10 |
| F5 | 7.557E-10 | 7.557E-10 | 7.557E-10 | 7.557E-10 | 7.557E-10 | 7.557E-10 | 7.557E-10 | 7.557E-10 |
| F6 | 7.550E-10 | 7.280E-10 | 7.557E-10 | 1.087E-09 | 7.550E-10 | 7.557E-10 | 2.113E-01 | 7.555E-10 |
| F7 | 7.557E-10 | 7.523E-10 | 7.557E-10 | 7.557E-10 | 7.557E-10 | 7.557E-10 | 7.557E-10 | 7.550E-10 |
| F8 | 7.557E-10 | 8.031E-10 | 7.557E-10 | 7.557E-10 | 7.557E-10 | 1.303E-09 | 8.728E-07 | 7.557E-10 |
| F9 | 7.850E-15 | 1.023E-14 | 4.660E-05 | 9.496E-07 | 7.850E-15 | 7.850E-15 | 7.850E-15 | 7.850E-15 |
| F10 | 7.850E-15 | 7.850E-15 | 7.797E-15 | 2.977E-15 | 2.977E-15 | 7.850E-15 | 7.850E-15 | 2.977E-15 |
| +/=/- | 10/0/0 | 10/0/0 | 9/1/0 | 7/3/0 | 10/0/0 | 10/0/0 | 9/1/0 | 10/0/0 |

# 三、模型

本研究提出了一种基于改进蛇群优化算法的 CNN-LSTM-attention-adaboost 混合模型框架，旨在实现高精度、高实时性的航班轨迹预测。该模型利用 adaboost 机制训练多个弱学习器，每个弱学习器通过卷积神经网络（CNN）提取输入数据的局部空间特征，同时利用长短期神经网络（LSTM）抽取数据的时间特征，并引入注意力机制动态强化关键时间步的建模能力，最终再将多个弱学习器组合成一个强大的集成模型。为进一步提升模型性能，提出了一种经过多项策略改进的蛇群优化算法来对模型的超参数进行寻优调整。实验表明，通过改进蛇群算法对 CNN-LSTM-attention-adaboost 模型的超参数进行动态优化，能够深度挖掘航迹数据中的时空特征，在本研究所使用的 VariFlight 航迹数据集上，模型预测结果的 MAPE 和 RMSE 分别降至 1.3504%和 125.4342，较传统的 CNN-LSTM 时序预测模型的预测误差降低了约 70%。本研究通过算法改进与模型架构的协同优化，为航班轨迹的预测提供了一种兼具强鲁棒性与高精度的方案。

如图 3 所示，本 CNN-LSTM-attention-adaboost 混合模型框架包含输入层、卷积层、池化层、LSTM 层、attention 层和输出层，其中：

（1）输入层：设计用于接收维度与筛选后数据特征相匹配的输入数据，时间步设置为 6。输入数据包含了经过筛选的航迹记录中的高度、经度、纬度、时间四维数据，并经过 MinMaxScaler 进行归一化处理，将数据缩放到 0 到 1 的范围内，以确保数据的一致性和模型训练的稳定性。这部分处理与代码中数据加载和归一化处理部分相对应，将原始数据处理成适合模型输入的格式。

（2）CNN 层：该层包含两个一维卷积层和一个最大池化层。卷积层使用 64 个大小为 3 的卷积核，步长默认为 1，激活函数采用 ReLU。其作用是通过卷积操作提取数据的局部特征，增强模型对输入数据中局部模式的敏感性。最大池化层的池化大小为 2，步长为 1，通过下采样操作降低数据维度，减少计算量的同时保留重要特征信息。利用蛇群优化算法逆向调整卷积核参数，探索更优的特征提取方式，有助于提高模型对局部特征的捕捉能力。

（3）LSTM 层：LSTM 层采用两层双向结构，第一层设 128 个神经元，第二层 64 个神经元，均配置 dropout 率为 0.2，利用 tanh 激活函数处理单元状态、sigmoid 函数控制门机制，第一层捕捉初级时序模式，第二层提取高级依赖关系，经蛇群算法优化神经元配比，结合双重 Dropout 抑制过拟合，最终输出完整序列供 Attention 层分配权重，形成能全面捕捉航迹数据复杂时序特征的深层网络结构。

（4）Atention 层：该层根据 LSTM 层的输出特征，为不同时间步的隐藏状态分配不同的权重。通过自定义的 AttentionLayer 实现，该层包含可训练的权重 W 和偏置 b，通过 K.tanh 和 K.softmax 函数计算注意力权重。权重值越大，表示对应时间步的特征对最终预测结果的影响越显著。通过这种方式，Attention 层能够突出对预测结果影响最大的关键特征，增强模型对关键时间步的捕捉能力。

（5）输出层：输出层由一个全连接层组成，包含 3 个神经元，对应高度、经度、纬度三个特征的预测值。该层未显式指定激活函数，使用默认的线性激活函数。最终输出的是经过反归一化处理的预测结果，即已有数据划分的验证集的高度、经度、纬度预测值。

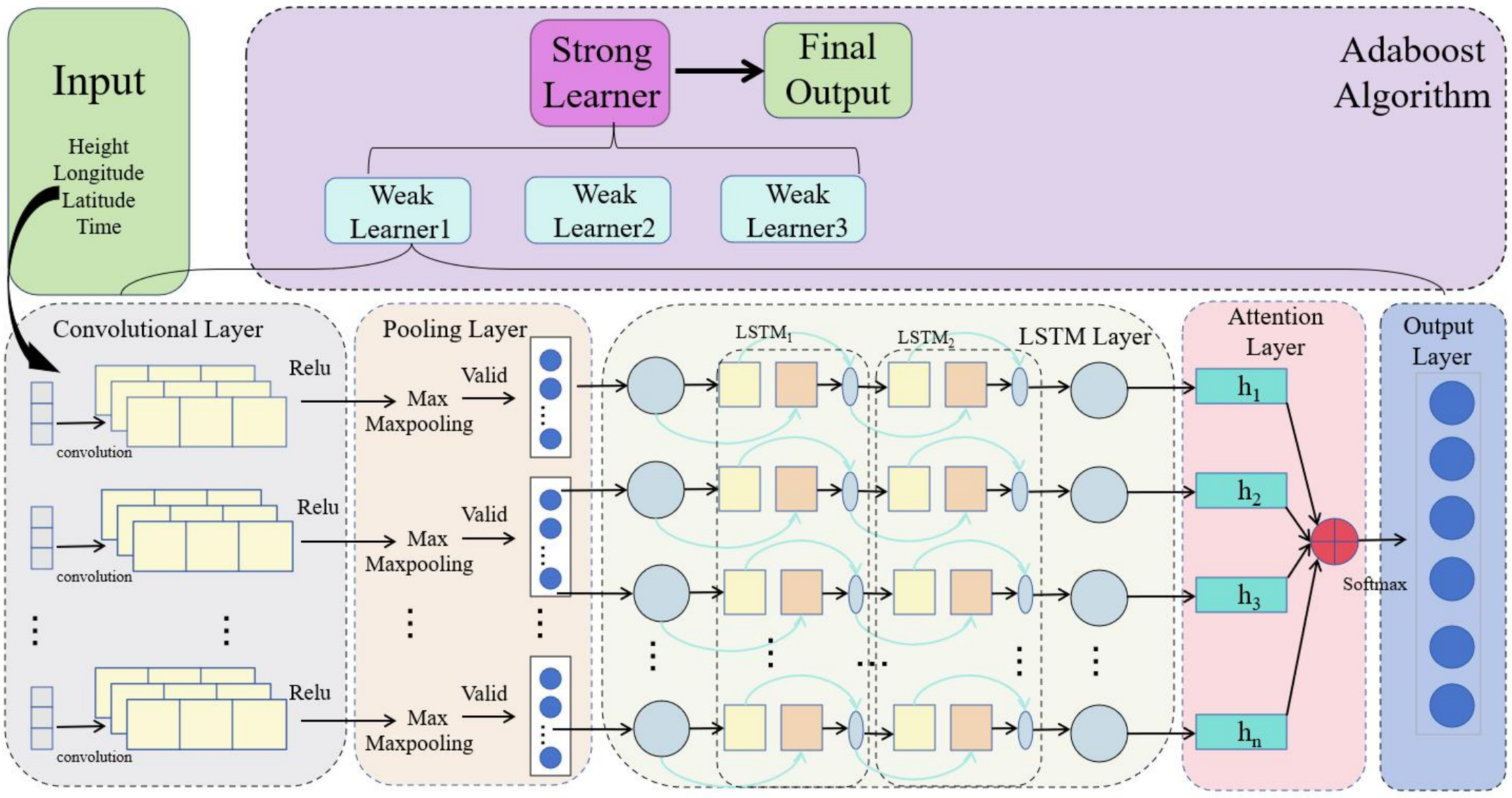

Figure4　CNN-LSTM attention adaboost hybrid model framework

## 3.1 卷积神经网络

卷积神经网络（CNN）是一种深度学习模型，专门用于处理网格结构数据（如图像或时间序列）。[28]卷积神经网络通过局部连接、权值共享和层次化特征提取机制，显著提升了图像等网格化数据的处理能力，解决了全连接网络参数量过大和平移不变性缺失的问题。其核心结构由卷积层、激活函数、池化层和全连接层构成。

在前向传播过程中，第 l 层卷积层的输入包括输入特征图、卷积核权重以及偏置项，其计算过程通过离散卷积运算实现：

$$Z^l_{(x,y,d)} = \sum_{C=1}^{C^{l-1}} \sum_{i=-k/2}^{k/2} \sum_{j=-k/2}^{k/2} W^l_{(i,j,c,d)} \cdot X^{l-1}_{(x+i,y+j,c)} + b^l_d$$

式中，$X^{l-1} \epsilon R^{H^{l-1}\times W^{l-1}\times C^{l-1}}$为输入特征图，$H^{l-1}$和$W^{l-1}$分别是输入特征图的高度和宽度，$C^{l-1}$为输入通道数；$W^l \epsilon R^{k\times k\times C^{l-1}\times C^l}$为卷积核权重，$k$为卷积核尺寸，$d$为输出通道索引；$b^l$为偏置项，$Z^l$为线性输出结果（未激活）。

每个卷积层的输出经过非线性激活函数（如 ReLU）变换后，进入池化层进行空间下采样，从而逐步构建起层次化的特征表示。池化层的引入进一步提升了网络的鲁棒性。最大池化操作通过在局部区域内选取最大值，实现了特征的下采样和位置不变性，公式如下：

$$P^l_{(x,y,c)} = max\, X^l_{(i,j,c)} \ ,\ i,j \,\epsilon\, R_{x,y}$$

式中，$R_{x,y}$为以(x,y)为中心的$k_p \times k_p$窗口（通常$k_p$=2）。

在网络末端，经过多次卷积和池化处理的高层特征会被展平并送入全连接层。这些全连接层的作用是将分布式特征表示映射到样本标记空间，完成最终的分类或回归任务。

$$\mathrm{y = \sigma(W^L vec(X^{L-1}) + b^L)}$$

式中，vec 将 $\mathrm{H \times W \times C}$ 特征图展平为向量，σ为 softmax 或 sigmoid。

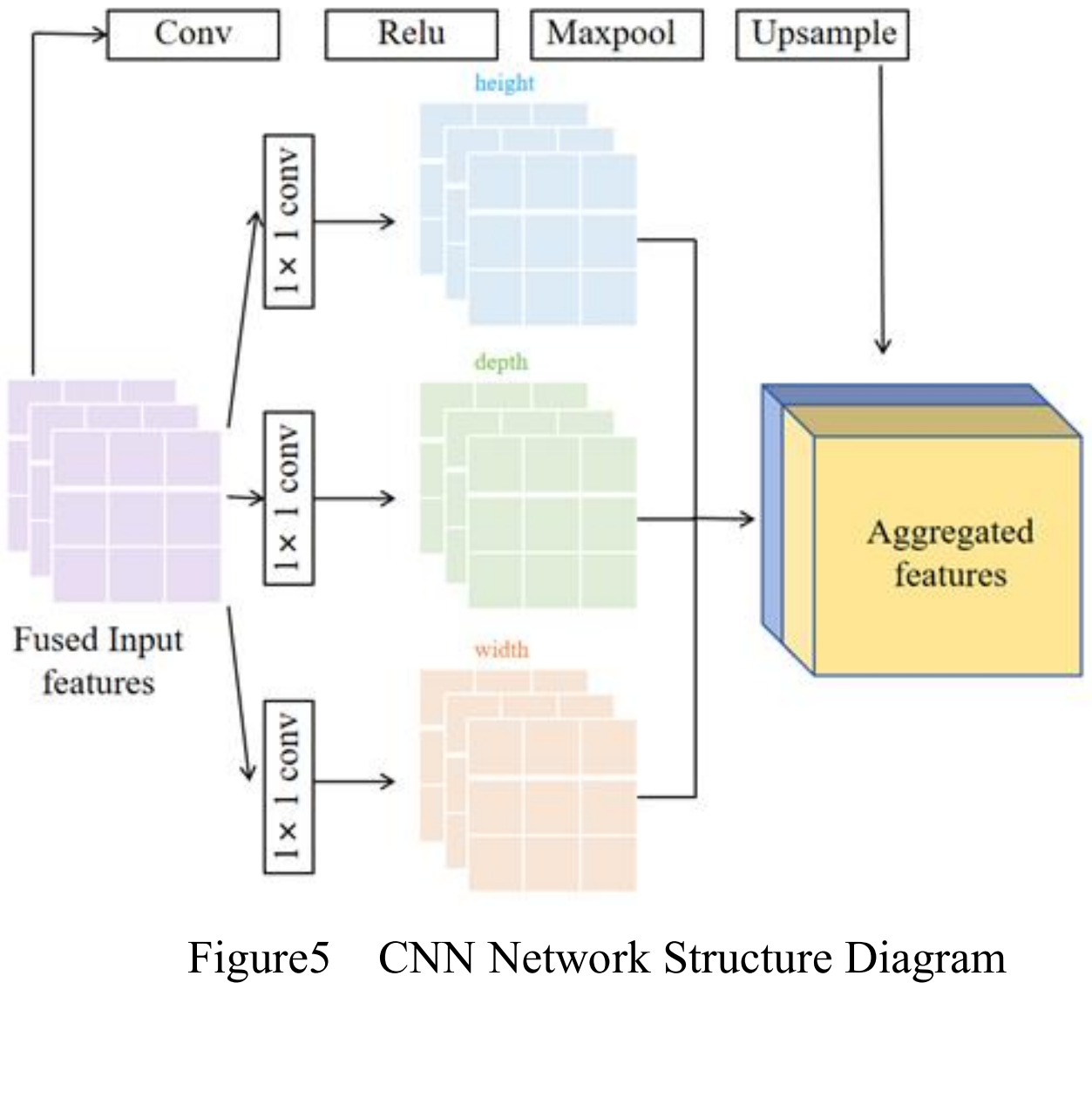

Figure5　CNN Network Structure Diagram

## 3.2 长短期神经网络

LSTM（长短期记忆网络）通过门控机制解决传统 RNN 梯度消失或梯度爆炸的问题，适用于处理长序列数据。[29]其核心结构包含遗忘门、输入门、输出门和细胞状态。

在 t 时刻，LSTM 当前的输入向量为$x_t$、上一时刻隐藏状态为$h_{t-1}$，细胞状态为$C_{t-1}$。

遗忘门决定上一时刻的细胞状态$C_{t-1}$中哪些信息被保留，计算公式如下：

$$f_t = \sigma(W_f[h_{t-1}, x_t] + b_f)$$

其中，$W_f$为遗忘门的权重矩阵，$b_f$为偏置向量，$\sigma$为$Sigmoid$激活函数，该函数将输出值压缩到[0,1]区间。$[h_{t-1}, x_t]$表示将上一时刻隐藏状态$h_{t-1}$与当前输入$x_t$进行拼接操作，输出的$f_t$中值越接近 1，表示对应信息保留程度越高；越接近 0，则表示信息被遗忘程度越高。

输入门负责确定当前时刻有哪些新的信息需要加入到细胞状态中，它由两部分组成：

$$i_t = \sigma(W_i[h_{t-1}, x_t] + b_i)\widetilde{C}_t = tanh(W_c[h_{t-1}, x_t] + b_c)$$

其中，$i_t$是输入门控信号，同样通过$Sigmoid$函数输出[0,1]区间的值，用于控制新信息的流入量；$\widetilde{C}_t$是候选细胞状态，通过$tanh$激活函数将值压缩到[-1,1]区间，代表当前时刻可能更新到细胞状态的新信息。

细胞状态更新是 LSTM 的关键步骤，它结合了遗忘门和输入门的信息：

$$C_t = f_t \odot C_{t-1} + i_t \odot \widetilde{C}_t$$

式中，$f_t \odot C_{t-1}$表示根据遗忘门的输出，选择性保留上一时刻细胞状态$C_{t-1}$中的信息；$i_t \odot \widetilde{C}_t$则根据输入门控信号$i_t$，将候选细胞状态$\widetilde{C}_t$中的部分新信息添加到细胞状态中。通过这样的机制，细胞状态能够在长序列中传递和更新信息，维持长期记忆。

输出门用于控制细胞状态中的信息如何输出到当前时刻的隐藏状态$h_t$：

$$o_t = \sigma(W_O[h_{t-1}, x_t] + b_O)h_t = o_t \odot tanh(C_t)$$

$o_t$是输出门控信号，由$Sigmoid$函数生成，用于调节细胞状态信息的输出比例；$tanh(C_t)$将细胞状态的值映射到[-1,1]区间，最终两者通过逐元素相乘操作$\odot$，得到当前时刻的隐藏状态$h_t$。$h_t$一方面会作为下一时刻 LSTM 单元的输入，另一方面也可作为模型的输出，用于后续的预测或分类任务。

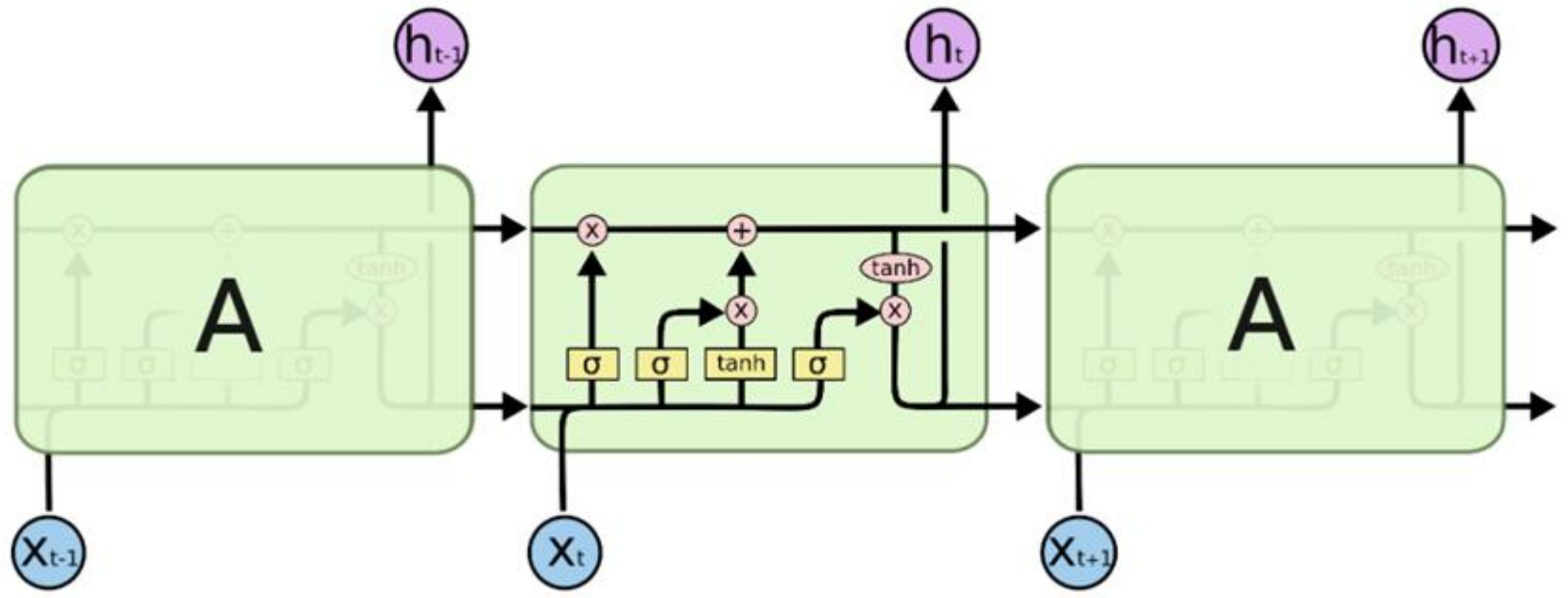

Figure6　LSTM Network Structure Diagram

## 3.3 注意力机制

随着深度学习的发展，模型在处理复杂任务时，需要对输入数据中的不同部分分配不同的关注程度。注意力机制（Attention）正是基于这一需求而被提出，它允许模型在计算过程中聚焦于输入的关键部分，动态地分配权重，从而更有效地提取重要信息。[30]在本模型中，LSTM 的输出序列会输入到注意力机制中，用于增强模型对重点时间步的关注度，注意力机制通过计算每个时间步的权重，决定哪些时间步对预测最重要，通过查询 Q，键 K，值 V

的关系来计算注意力分数。

注意力权重计算如下：

$$\alpha_t = softmax(\frac{QK^T}{\sqrt{d_k}})$$

加权输出：

$$\mathrm{E(t)} = \sum_{t'} \alpha_{t'} \mathrm{V}(t')$$

式中，其中$d_k$为键的维度，V(t')是值矩阵，代表每个时间步的隐藏状态，$\alpha_t$表示每个时间步的注意力权重，计算加权隐藏状态的值，E(t)表示对所有时间步的加权和，聚焦于重要的时间步。

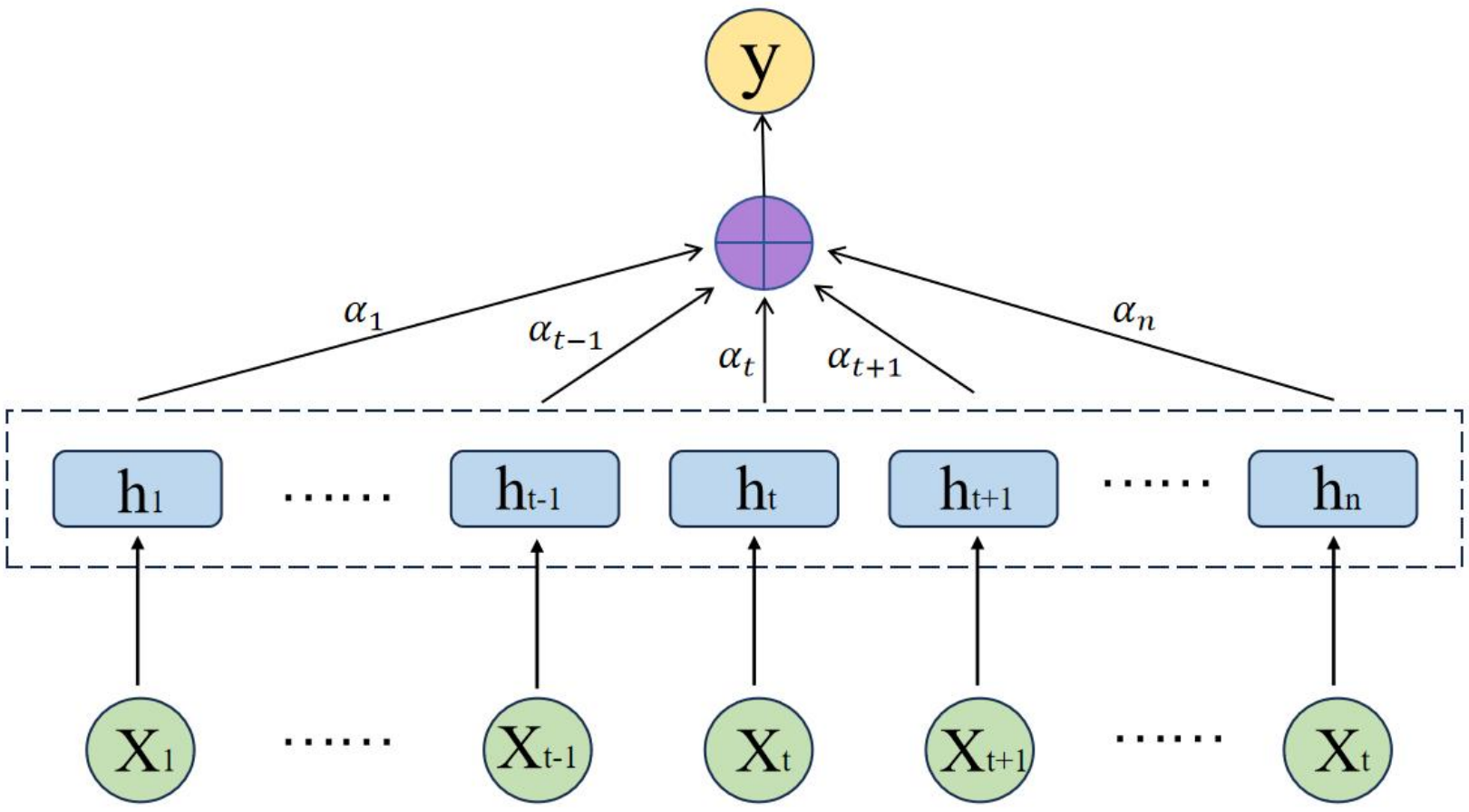

Figure7　Attention Mechanism Structure Diagram

## 3.4　Adaboost 算法

Adaboost 作为经典的集成学习算法，通过迭代训练多个弱分类器并加权组合成强分类器，有效提升模型的分类性能，克服单一弱分类器精度不足的问题。[31]其核心在于通过自适应调整样本权重和弱分类器权重，逐步聚焦于难分样本，优化整体分类效果。

给定训练数据集：

$$\mathrm{T} = \{(x_1, y_1), (x_2, y_2), \cdots\cdots, (x_N, y_N)\}$$

式中，$x_i \in \chi \subseteq R^n$为输入样本，$y_i \in \gamma = \{+1, -1\}$为类别标签。

算法初始时赋予所有样本等权重分布：

$$D_1(i) = \frac{1}{N}$$

样本权重初始化后，进入迭代训练阶段。在第 m 次迭代中，基于当前样本权重分布$D_m$训练一个弱分类器$G_m(x)$。随后，通过以下公式计算该弱分类器在当前权重下的错误率：

$$e_m = \sum_{i=1}^{N} D_m(i) I\left(G_m(x_i) \neq y_i\right)$$

其中，$I(\cdot)$为指示函数，当括号内条件成立时取值为 1，否则为 0；$D_m(i)$表示第$m$次迭代时第$i$个样本的权重。该公式通过加权求和，量化了弱分类器对样本的分类错误程度。根据计算所得的错误率$e_m$，利用公式确定该弱分类器的权重：

$$\alpha_m = \frac{1}{2} ln(\frac{1-e_m}{e_m})$$

此公式表明，弱分类器的错误率$e_m$越低，$\alpha_m$越大，意味着该弱分类器在最终强分类器中所占权重越高，对分类结果的影响越大。

Adaboost 通过以下公式更新样本权重：

$$D_{m+1}(i) = \frac{D_m(i)}{Z_m} exp(-\alpha_m y_i G_m(x_i))$$

其中，$Z_m = \sum_{i=1}^{N} D_m(i) exp(-\alpha_m y_i G_m(x_i))$为归一化因子，确保$D_{m+1}$构成概率分布。在该公式中，若样本被正确分类，$y_i G_m(x_i)$为正，样本权重降低；若被错误分类，$y_i G_m(x_i)$为负，样本权重升高，从而实现对难分样本的重点关注。

经过 M 次迭代后，将所有弱分类器按照权重累加，通过符号函数输出最终的强分类器：

$$G(x) = sign(\sum_{m=1}^{M} \alpha_m G_m(x))$$

该公式通过对加权后的弱分类器输出进行符号判断，得出最终分类结果，完成了从多个弱分类器到强分类器的转变。

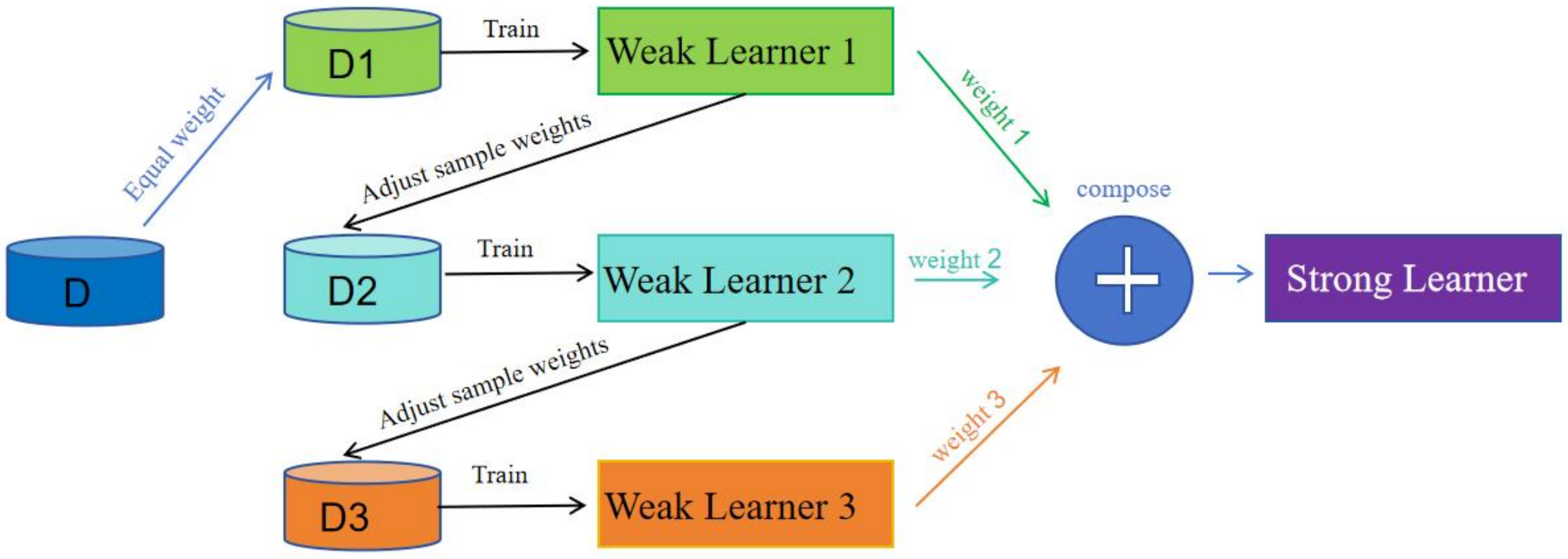

Figure8 Structure diagram of adaboost algorithm

# 四、实验分析与结果

## 4.1 数据来源和模型评价指标选取

本文使用的数据集来源于天津至西安航线某架航班的 ADS-B 记录数据，数据来源于 VariFlight 技术公司的数据( https://flightadsb. variflight. com/track-data)。[32]数据收集时间范围为 2024 年 4 月至 2024 年 7 月，经过筛选后共计获得 20526 条航迹记录，每条航迹记录包含飞行器在不同时间点的经纬度、高度、速度、航向等重要特征信息。输入数据为已有数据划分的训练集的高度、经度、纬度真实值，输出数据为已有数据划分的验证集的高度、经度、纬度预测值。

为评估本文模型的预测精度，选用平均绝对百分比误差(MAPE)、均方根误差(RMSE)、判定系数($R^2$)、平均绝对误差（MAE）和最大绝对误差(MAXAE)作为模型预测结果的评价指标。[33]其中：MAPE 和 RMSE 用来衡量预测精度，二者值越小则预测精度越高；$R^2 \in [0,1]$ 用来反映模型优劣， $R^2$ 越接近 1 则预测模型性能越好；MAE 计算的是预测值与真实值误差绝对值的平均值，直观体现平均偏差程度；MAXAE 关注的是在给定样本范围内预测值与真实值误差绝对值中的最大值，突出极端情况下的偏差情况。

## 4.2 不同智能算法优化 CNN-LSTM-attention-adaboost 模型超参数的预测结果对比

本研究用 SO 算法与八种对比算法分别对 CNN-LSTM-attention-adaboost 组合模型的超参数进行优化，以此来观察并对比优化后模型的预测性能。[34]各算法初始的参数设置如表几所示，具体的寻优过程与寻优分布分别如图几和图几所示。批大小、学习率和神经元数量三个超参数的寻优范围分别设置为[16，128]、[0.0001，0.02]、[50，200]。

表几为九种算法各自寻优得到的学习率、神经元数量、批大小和最优损失值，从中可以看到，SO 算法具有其中最快的寻优速度，无论是批大小、学习率还是神经元数量均能在 16 代以内寻到最优，并且最终损失值为 0.000544，是九种算法中最小的，这表明了 SO 拥有着相较于其他算法更高的寻优效率与优化效果；WOA 算法的最终损失值为 0.000801，批大小寻优较快，在 20 代左右便寻至最优，但神经元数量寻优较慢，在 38 代才到达最优；GWO 算法的最终损失值为 0.000855，是九种算法中最差的，这可以归咎于超参数寻优速度过慢，三个超参数均在 26 代之后才达到最优，其中学习率更是在第 43 代才寻到最优解，没有为模型留下充足收敛的时间；BSO 算法与 PSO 算法的情况相仿，所有超参数均能在 30 代以内寻至最优，且超参数数值跳变幅度较小，展现出了良好的稳定性；DE 算法的最终损失值为 0.000721，其中，神经元数量的寻优在早期出现了反复性的剧烈跳变，从第 10 代之后便逐渐趋于稳定；GA 算法的最终损失值为 0.000706，神经元数量同样是在寻优早期出现了剧烈的跳变，随后便呈现了阶梯性的持续走低；BBO 算法的最终损失值为 0.000759，学习率与神经元数量寻优较快，都在 20 代左右便寻到最优值，但批大小直到第 37 代才稳定下来；

GCO 算法的最终损失值为 0.000626，三个超参数寻优效率均较高，分别在第 18、19、29 代寻得最优值，并且所有超参数的跳变幅度极小，算法稳定性极佳。

Table4　Initial parameter settings for nine optimization algorithms

| WOA | 参数设定 | GWO | 参数设定 | BSO | 参数设定 |
|---|---|---|---|---|---|
| 鲸鱼数量 | 30 | 狼群规模 | 30 | 种群大小 | 30 |
| 收缩包围因子 | 初始为 2，线性递减 | 位置更新系数 | 初始为 2，线性递减 | 学习因子 | 0.1 |
| 距离控制系数 | 0.5 | 探索系数 | [0,1]随机数 | 变异概率 | 0.05 |
| 初始能量值 | 10 | 开发系数 | [0,1]随机数 | 邻域大小 | 50% |

| PSO | 超参数设定 | DE | 参数设定 | GA | 参数设定 |
|---|---|---|---|---|---|
| 种群大小 | 30 | 种群大小 | 50 | 种群大小 | 10 |
| 惯性权重 | 0.5 | 变异因子 | 0.5 | 交叉概率 | 0.5 |
| 个体学习因子 | 1.5 | 交叉概率 | 0.3 | 变异概率 | 0.2 |
| 社会学习因子 | 1.5 | 变异策略 | DE/rand/1 | 交叉方式 | 线性交叉 |
| | | | | 变异方式 | 高斯变异 |

| BBO | 参数设定 | GCO | 参数设定 | SO | 参数设定 |
|---|---|---|---|---|---|
| 种群大小 | 20 | 种群大小 | 15 | 种群大小 | 30 |
| 迁移率 | 0.5 | 惯性权重 | 0.6 | 探索系数 | 0.5 |
| 变异率 | 0.05 | 学习因子 | 1.2 | 开发系数 | 2 |
| 最大迁移距离 | 10 | 警觉性因子 | 0.5 | 食物阈值 | 0.25 |
| | | | | 温度阈值 | 0.6 |

Table5　Optimization results of nine algorithms

| | Stabilizing Iteration | Loss | Node | Batch size | Learning rate |
|---|---|---|---|---|---|
| WOA | 38 | 0.000801 | 134 | 27 | 0.0086 |
| GWO | 43 | 0.000855 | 142 | 84 | 0.0052 |
| BSO | 29 | 0.000689 | 102 | 35 | 0.0068 |
| PSO | 27 | 0.000635 | 58 | 84 | 0.0064 |
| DE | 35 | 0.000721 | 140 | 39 | 0.0044 |
| GA | 31 | 0.000706 | 145 | 36 | 0.0072 |
| BBO | 37 | 0.000759 | 158 | 38 | 0.0071 |
| GCO | 30 | 0.000626 | 106 | 44 | 0.0077 |
| SO | 16 | 0.000544 | 71 | 48 | 0.0076 |

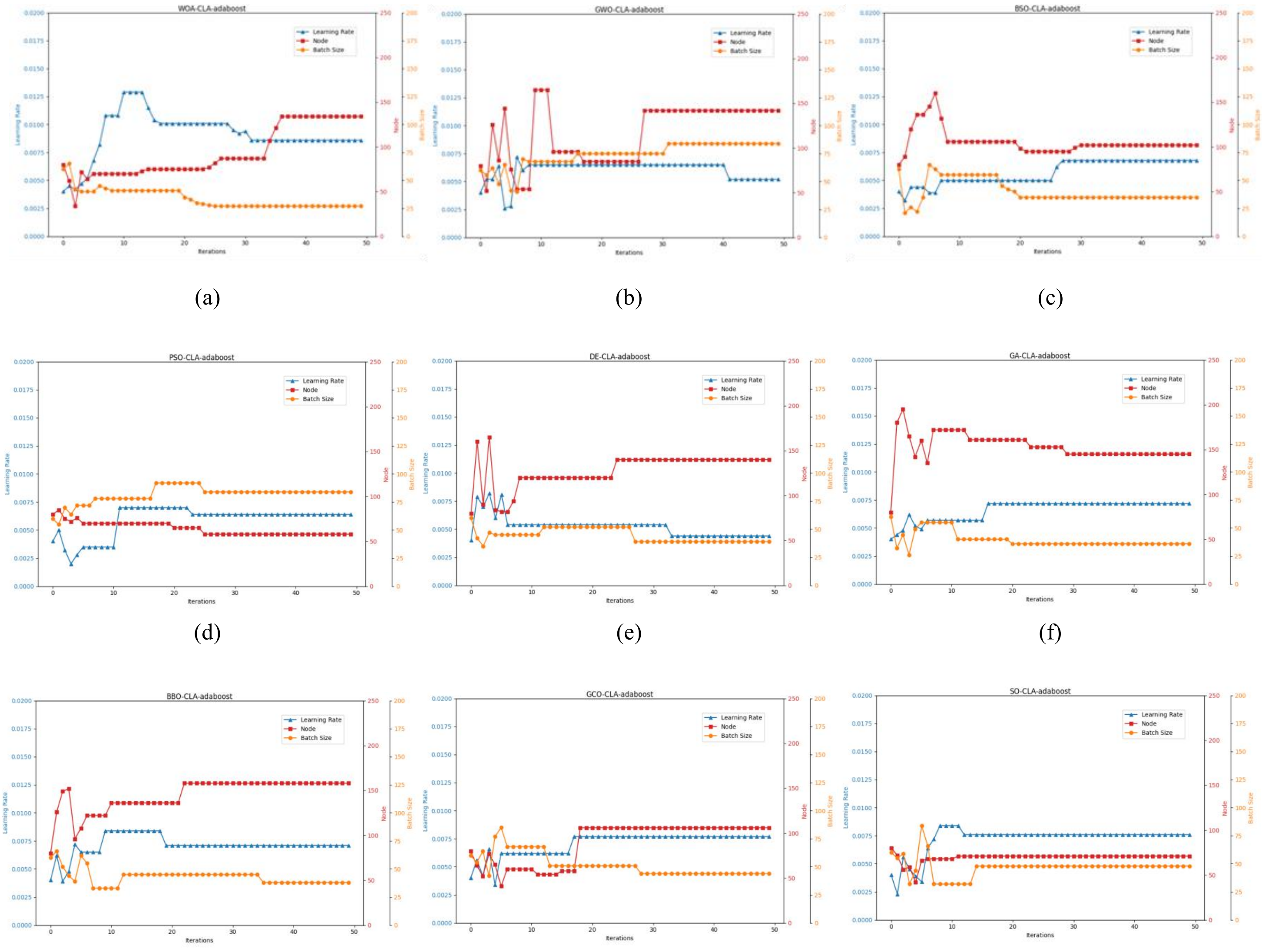

(a) (b) (c)

(d) (e) (f)

(g) (h) (i)

Figure9 Model hyperparameter optimization process

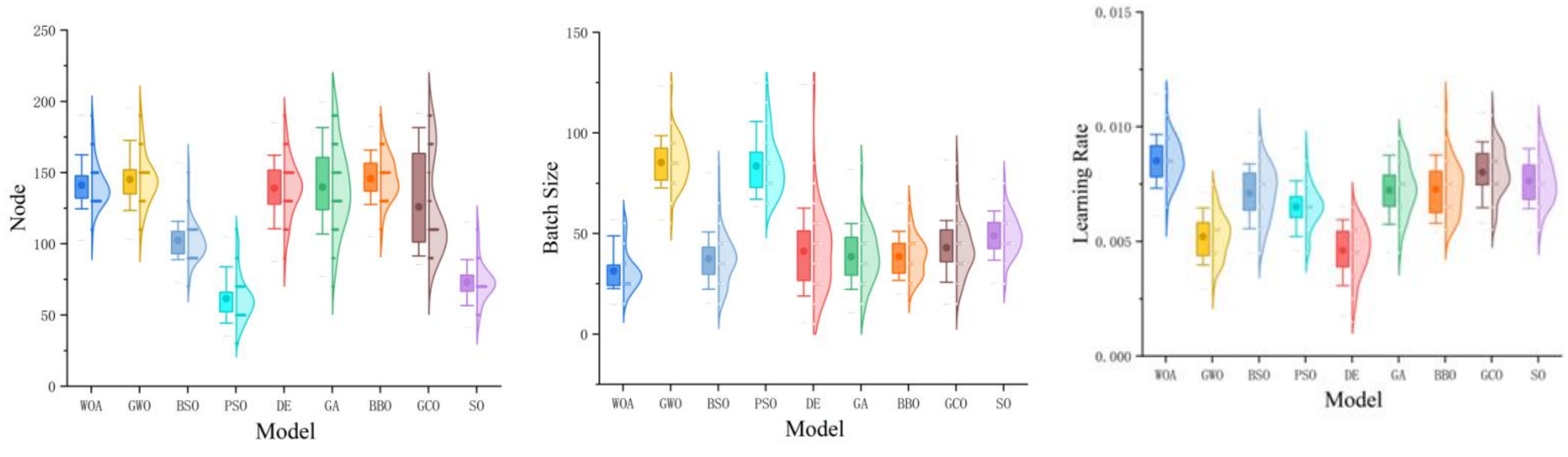

(a) (b) (c)

Figure10 Model hyperparameter optimization distribution

表几为本研究所构筑的不同模型在航班数据集上的预测性能对比。

在 RMSE 指标上，SO-CLA-adaboost 模型已经降至了 125.4342，为所有算法中最低的，仅有 WOA-CLA-adaboost 和 PSO-CLA-adaboost 能在该项指标上接近，但 SO-CLA-adaboost 仍能比二者分别降低 19.44%和 20.74%，而对比误差最大的 DE-CLA-adaboost 模型，足足降低了 56.94%。

在 MAPE 指标上，SO-CLA-adaboost 模型为 1.3504%，同样也是所有算法中最低的，相较于表现次之的 WOA-CLA-adaboost 与 GWO-CLA-adaboost，分别降低了 29.89%和 40.68%，与在该指标上表现最差的 GCO-CLA-adaboost 对比，降低了 56.97%。

在 MAE 指标上，SO-CLA-adaboost 模型已经能控制在 50 左右，而其他模型中表现最好的 WOA-CLA-adaboost 也只是保持在 64.0093，而表现最差的 DE-CLA-adaboost 更是高达 121.8376。

在 $R^2$ 指标上，SO-CLA-adaboost 模型达到了 0.9978，次之的 WOA-CLA-adaboost 和 PSO-CLA-adaboost 仅仅达到了 0.9962 和 0.9959，相较于 SO-CLA-adaboost 存在较大的差距，而在该项指标上表现最差的 DE-CLA-adaboost 堪堪突破 0.99。

在 MAXAE 指标上，各模型均存在较大的误差，这是因为航班在实际飞行中往往存在短时间内的海拔急剧爬升与下降，而模型无法瞬间同步地捕捉到这一趋势，因此模型预测的航路轨迹往往是相较于真实航迹慢一拍的。从表格中可以看到，其他算法中表现最好的 WOA-CLA-adaboost 模型在这些海拔剧变点上也存在着最大误差值到达 1274.5740 的海拔误差，GA-CLA-adaboost 模型的最大海拔误差更是高达 2042.1303，而 SO-CLA-adaboost 模型在这一指标上实现了跨越性的突破，将最大误差值降至 609.4213，相较于 WOA-CLA-adaboost 与 GA-CLA-adaboost 分别降低了 52.19%和 70.16%。

Table6　Quantitative indicators of model prediction performance

| index | RMSE | MAPE | MAE | MAXAE | $R^2$ | Rank |
|---|---|---|---|---|---|---|
| WOA-CLA-adaboost | 155.7012 | 1.9262% | 64.0093 | 1274.5740 | 0.9962 | 2 |
| GWO-CLA-adaboost | 178.6107 | 2.2764% | 75.0152 | 1523.8170 | 0.9953 | 4 |
| BSO-CLA-adaboost | 188.3953 | 2.5474% | 101.4109 | 1677.7378 | 0.9937 | 6 |
| PSO-CLA-adaboost | 158.2536 | 2.5412% | 91.4552 | 1273.3192 | 0.9959 | 3 |
| DE-CLA-adaboost | 291.3221 | 3.0698% | 121.8376 | 1923.1047 | 0.9907 | 9 |
| GA-CLA-adaboost | 231.1606 | 2.4725% | 86.2341 | 2042.1303 | 0.9938 | 8 |
| BBO-CLA-adaboost | 180.2478 | 2.6286% | 79.2887 | 1490.8533 | 0.9958 | 5 |
| GCO-CLA-adaboost | 222.8798 | 3.1385% | 103.3800 | 1782.8380 | 0.9913 | 7 |
| SO-CLA-adaboost | 125.4342 | 1.3504% | 51.9559 | 609.4213 | 0.9978 | 1 |

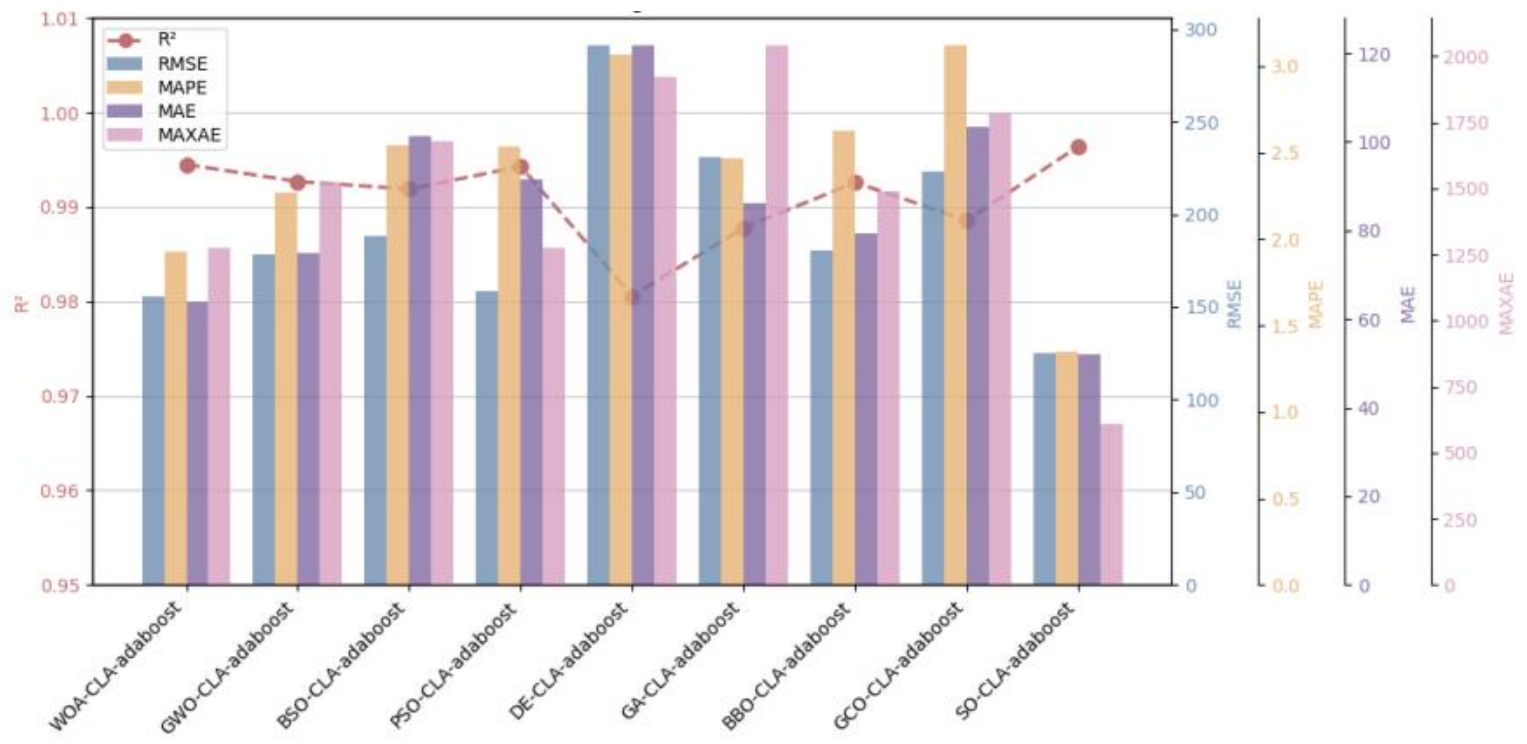

Figure11　Quantitative assessment analysis

## 4.3 本模型的预测值和真实值的对比分析

在高度维度上，各模型的表现差距不大，都能较为精准地预测出与真实值基本吻合的结果，其中 SO-CLA-adaboost 表现最好，几乎与真实航迹完全重合，GCO-CLA-adaboost 和 BBO-CLA-adaboost 表现最差。

在经度维度上，各模型在攀升及巡航阶段都基本能较完美地预测出准确的航迹，而在降落阶段则出现了较大的差异，所有算法都出现了锯齿状的波动，其中表现较好的是 SO-CLA-adaboost、BBO-CLA-adaboost、GCO-CLA-adaboost 和 GA-CLA-adaboost。SO-CLA-adaboost 在降落阶段的初期产生了较大的波动，后续便迅速地消失并与真实航迹贴合；GCO-CLA-adaboost 和 GA-CLA-adaboost 持续锯齿状波动但幅度逐渐减小；而 BBO-CLA-adaboost 在降落后期波动逐渐减缓至平滑，但依旧存在较小的误差。

在纬度维度上，除了 GCO-CLA-adaboost 出现了明显偏差以外，其他模型在攀升及巡航阶段都能较好地预测出准确的航迹；进入降落阶段后，除了 SO-CLA-adaboost 还能较好地贴近真实航迹之外，其他模型都产生了持续性的波动，其中以 BSO-CLA-adaboost 和 GA-CLA-adaboost 的表现最差。

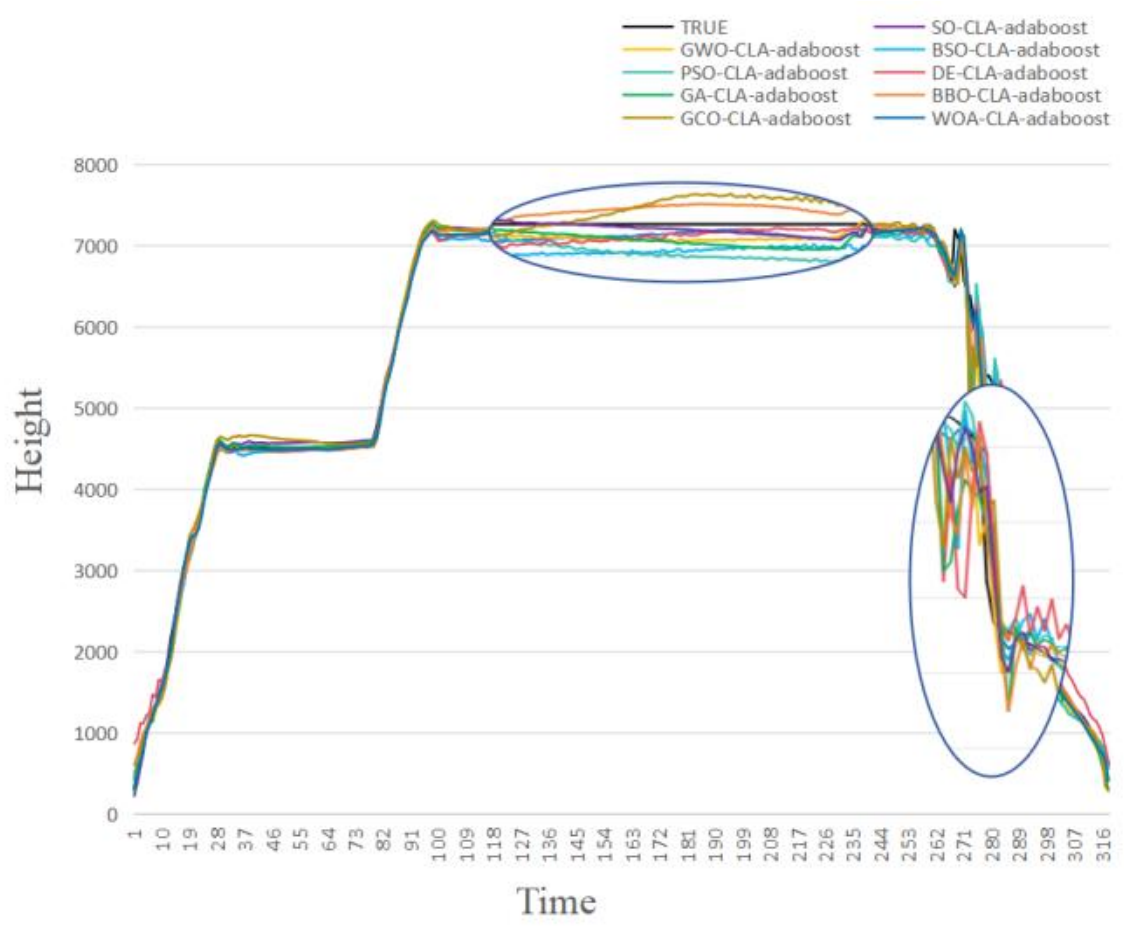

(a)

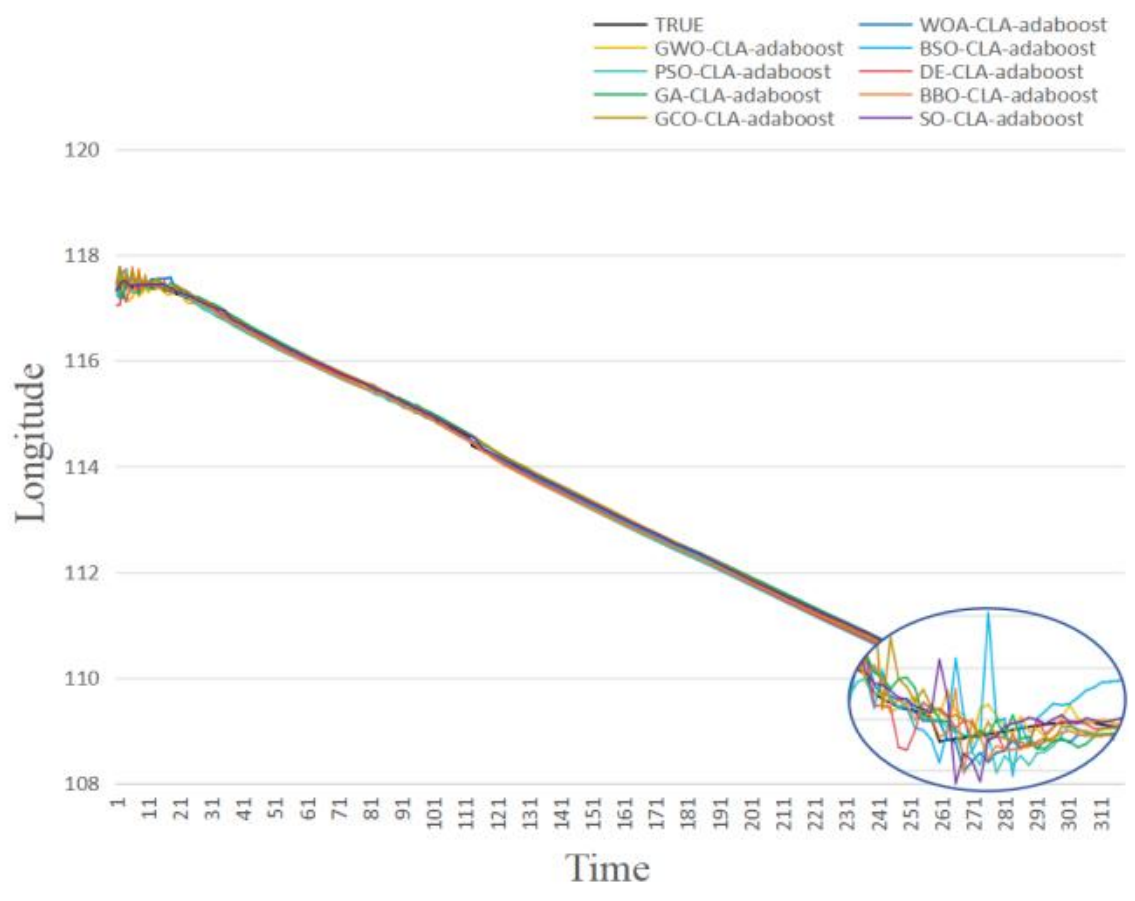

(b)

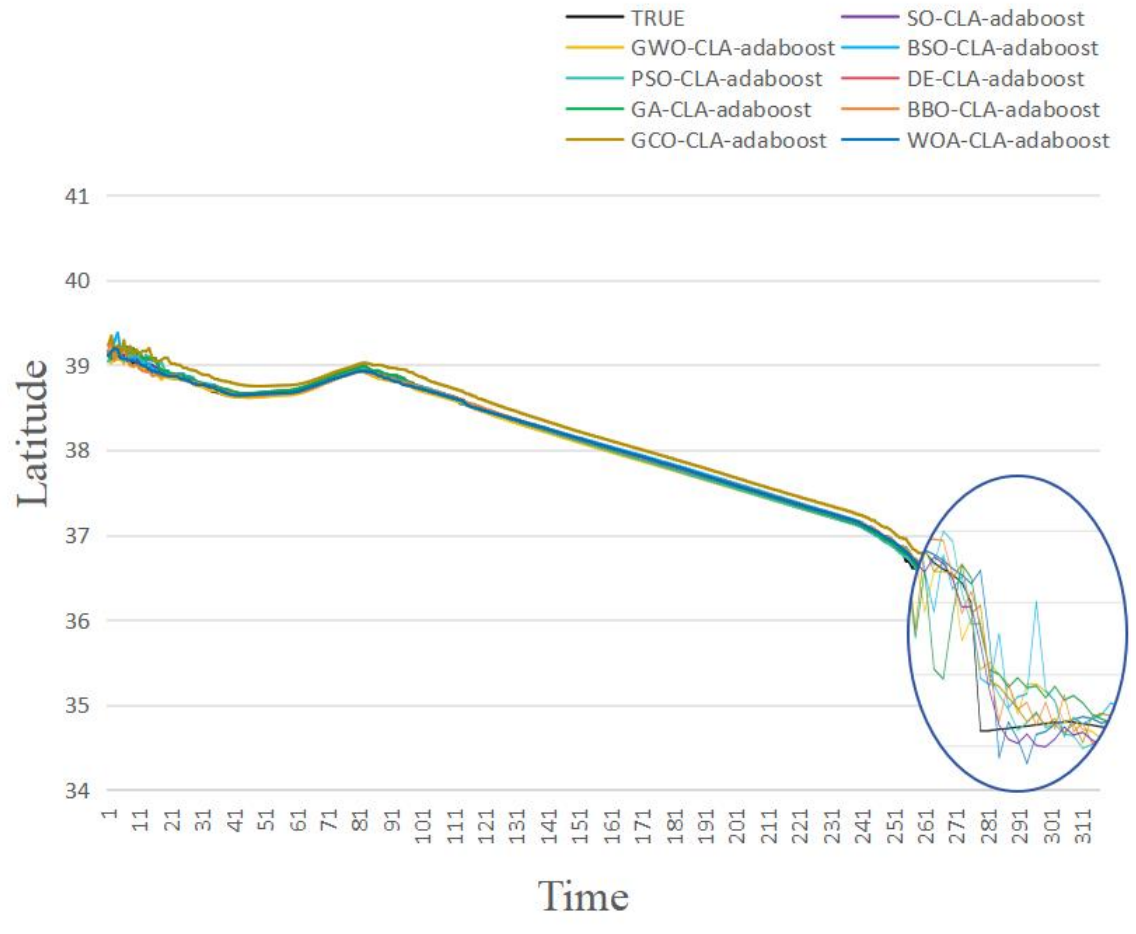

(c)

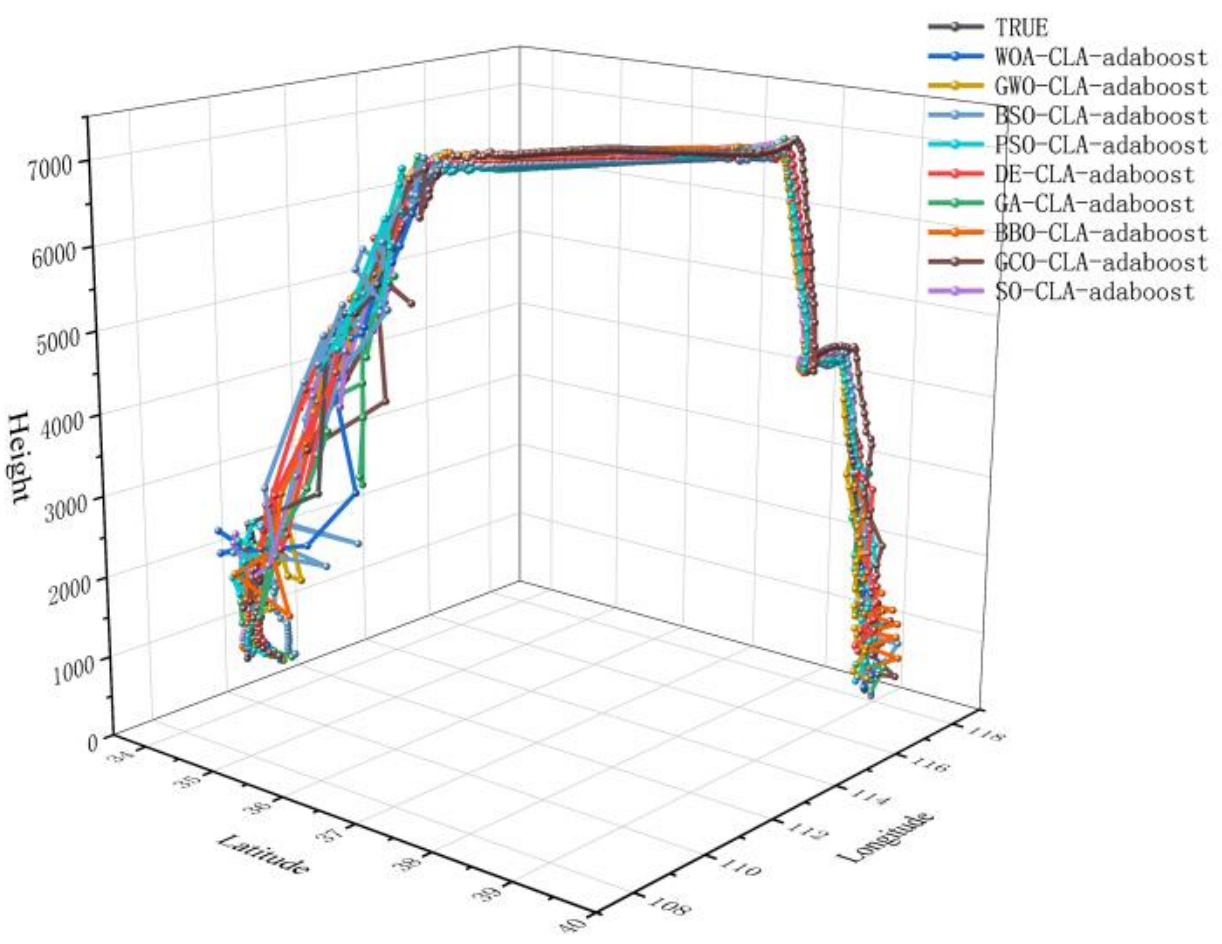

(d)

Figure12 Comparison of 3D trajectory prediction and individual comparison of different dimensions

对于各模型的三维航迹预测效果，本节着重针对攀升阶段和降落阶段进行了分析：

在攀升阶段，各模型在 0-4000m 的持续爬升飞行预测中差距较大，其中 SO-CLA-adaboost 的预测效果最好，仅在 1000m 高度以下出现了较大的误差，其他高度段均较精准地实现了飞行轨迹的预测；WOA-CLA-adaboost 和 GWO-CLA-adaboost 的预测效果其次，其中 WOA-CLA-adaboost 能较准确地把握航迹变化的趋势，但在预测精度上略有欠缺，存在较为明显的误差，而 GWO-CLA-adaboost 的预测精度不错，但没能准确捕捉到变化趋势，在上升过程中出现了持续性的小幅波动；BSO-CLA-adaboost、BBO-CLA-adaboost 和 GA-CLA-adaboost 在上升初期误差较大，出现了锯齿状的轨迹波动，但随着高度的增加逐渐减小；PSO-CLA-adaboost 和 DE-CLA-adaboost 的表现类似，在上升的中期逐渐出现了大幅的波动；GCO-CLA-adaboost 的预测表现最差，在 0-4000m 的爬升进程中都与真实航迹产生了较大的偏差。

在高度 4000-5000m 的盘旋飞行中其他模型均表现良好，基本能与真实航迹较好的吻合，仅 DE-CLA-adaboost 和 GCO-CLA-adaboost 出现了明显的预测误差。

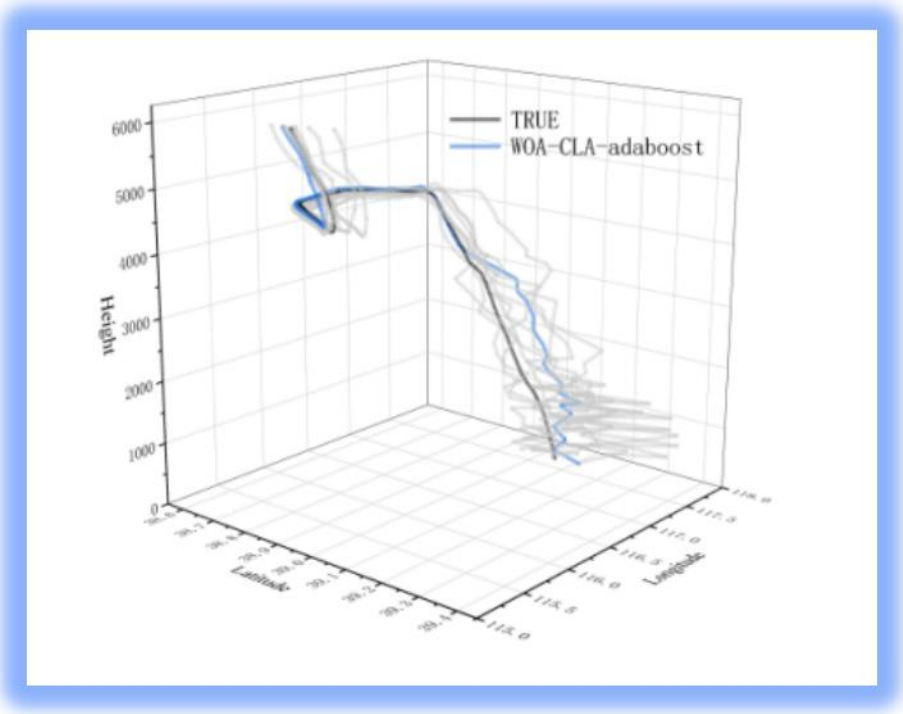

(a)

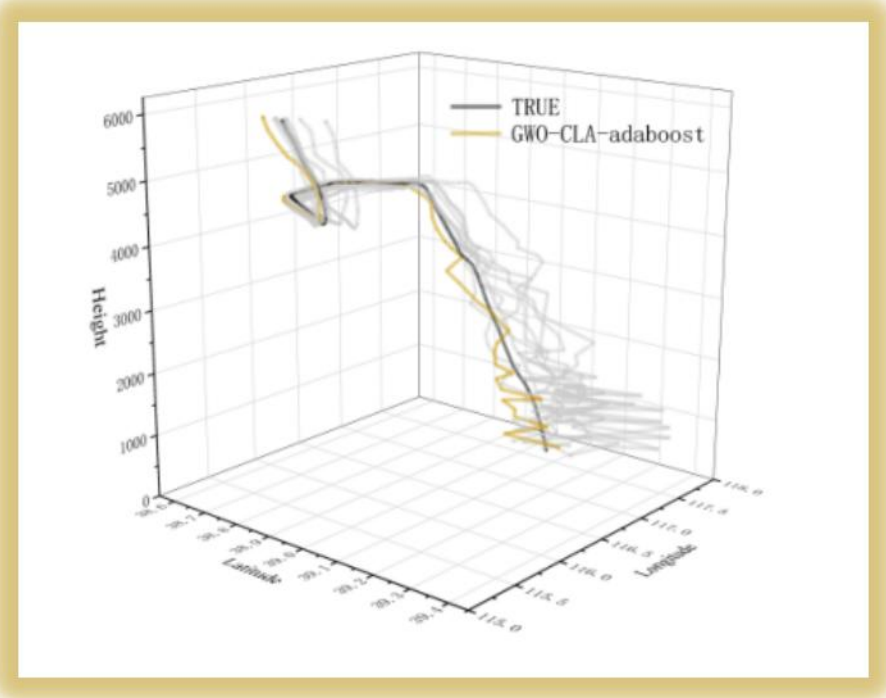

(b)

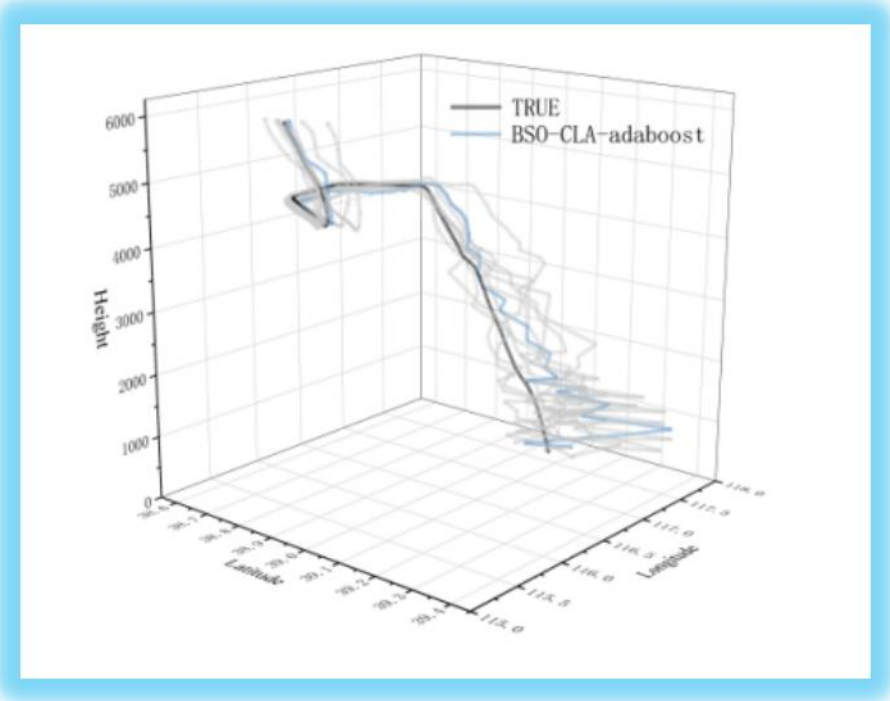

(c)

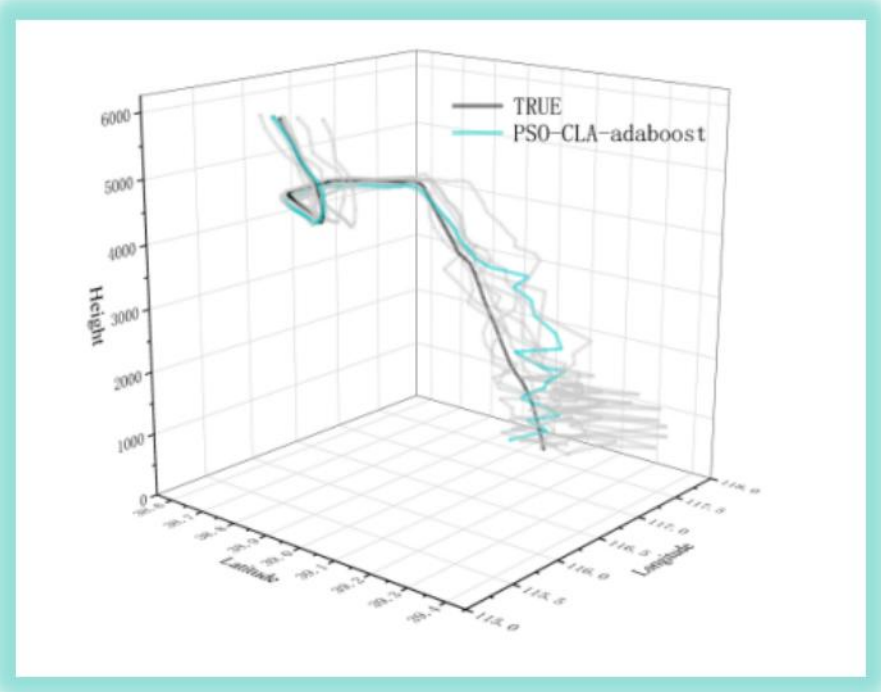

(d)

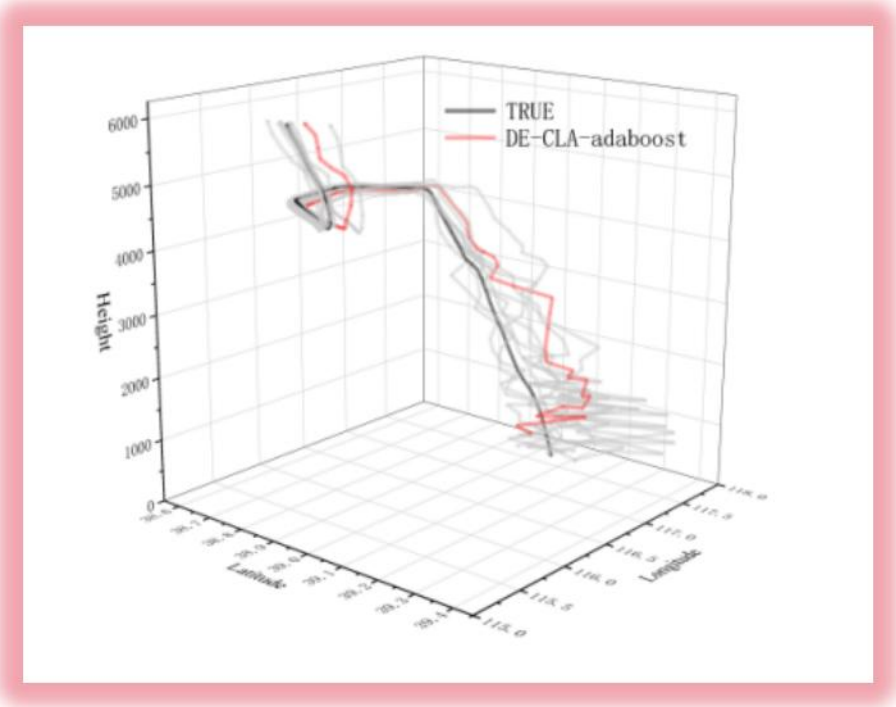

(e)

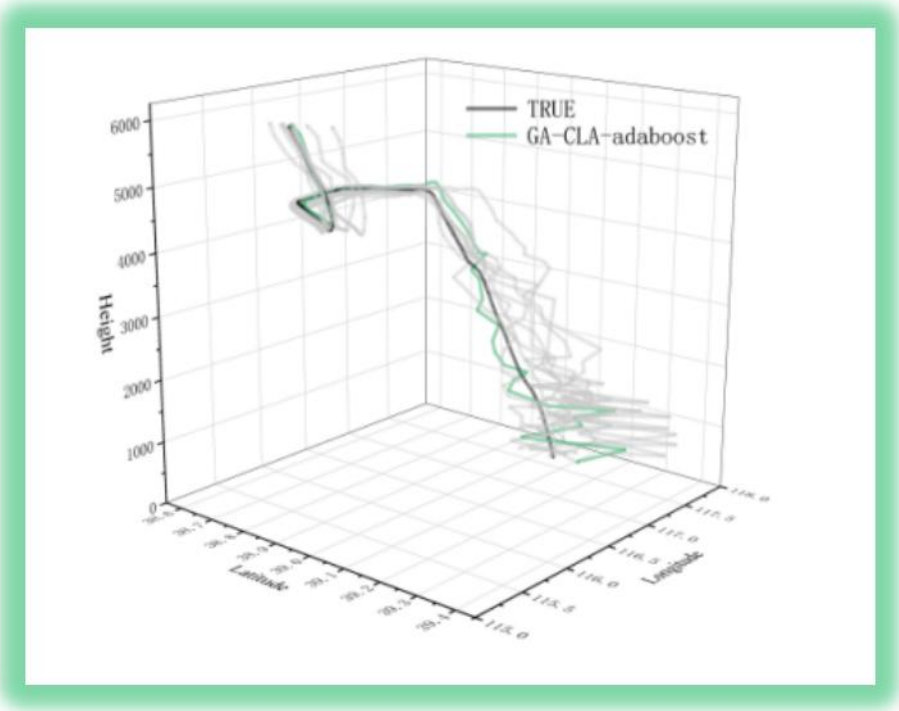

(f)

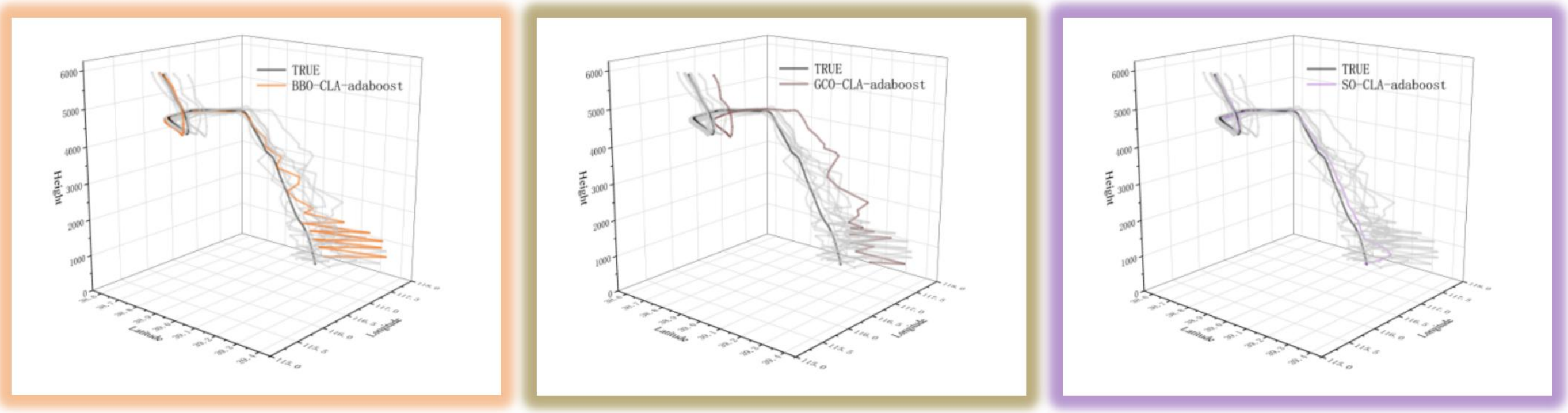

(g) (h) (i)

Figure13 Comparison between actual values and predicted values of various models during the climbing phase

在降落阶段，受限于降落时高度的骤降时机较难捕捉，相较于攀升阶段飞行高度的平稳上升，各模型的预测表现都不是非常理想。SO-CLA-adaboost 在下降初期，即 7000-4500m 的降落过程中预测效果相当出色，较为完美地吻合了真实航迹，但没能准确把握住 4000-2000m 的骤然下降，出现了较明显的误差；WOA-CLA-adaboost、GWO-CLA-adaboost、PSO-CLA-adaboost 和 BBO-CLA-adaboost 的预测表现相仿，都没有出现很大的误差，但在轨迹的变化趋势上已经无法较为精准的把握；BSO-CLA-adaboost 和 GA-CLA-adaboost 会分别在下降的某个阶段会出现大幅度的跳变；而 DE-CLA-adaboost 和 GCO-CLA-adaboost 预测出的航迹分别在 5000-2500m 和 6000-3000m 的高度段出现了持续性的往复变化，这意味着短时间内已经失去了对真实航迹预测的准确性。

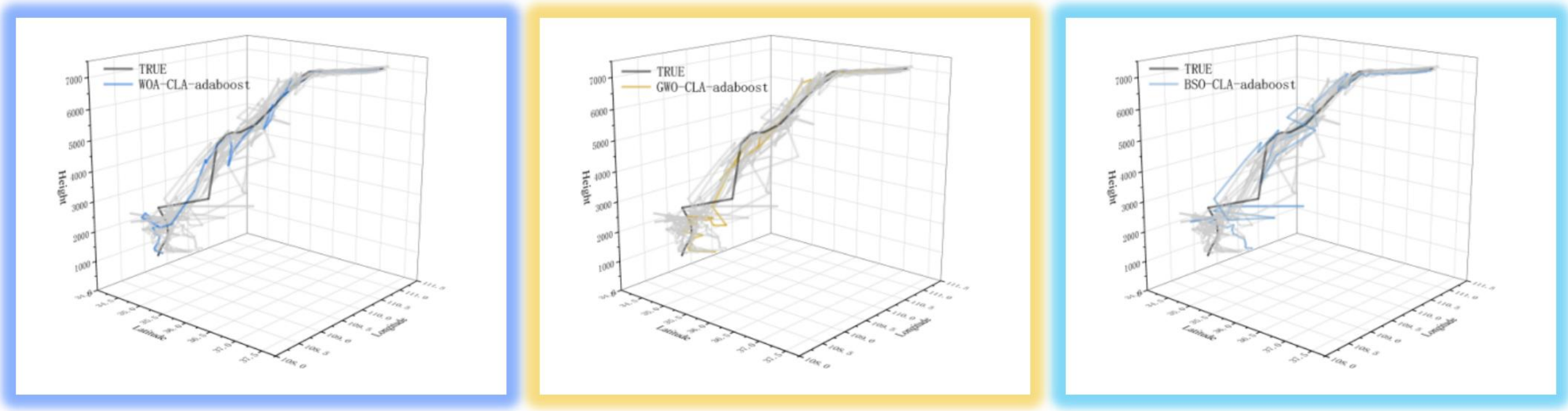

(a) (b) (c)

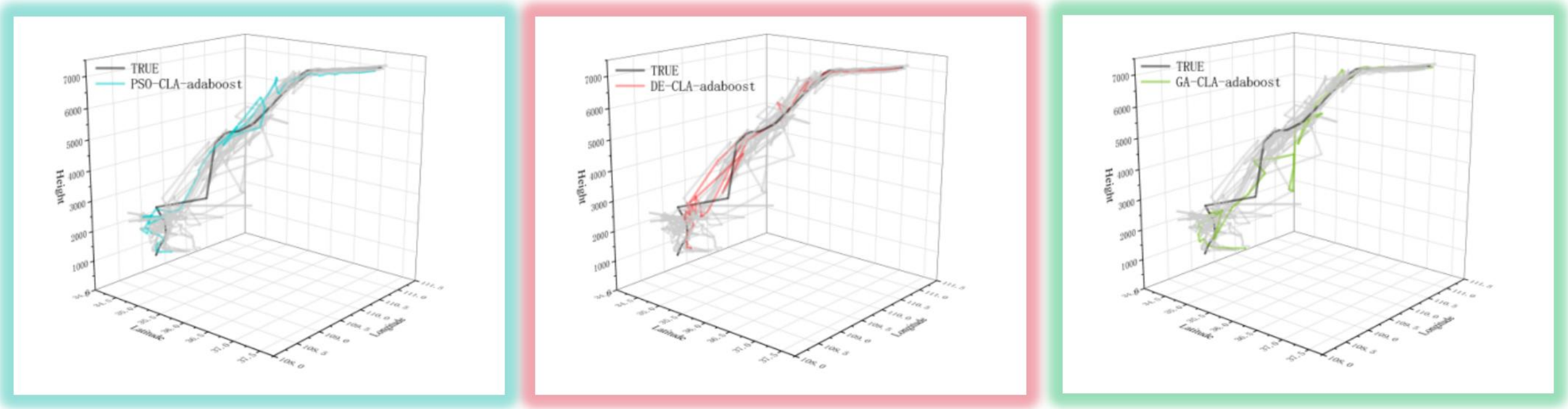

(d) (e) (f)

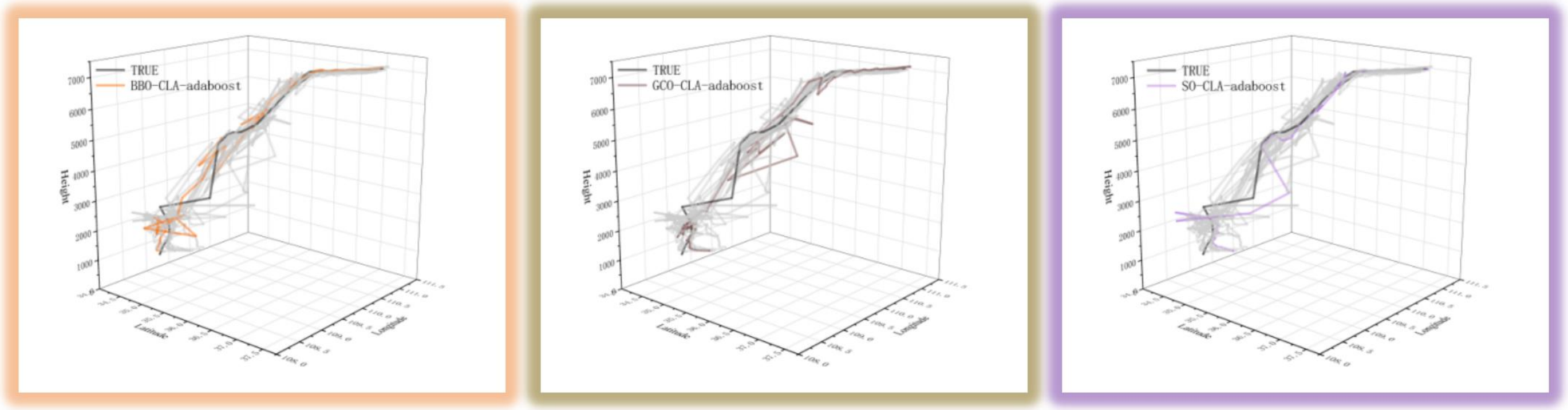

(g) (h) (i)

Figure14 Comparison between predicted values and actual values of various models during the landing phase

## 4.4 消融实验

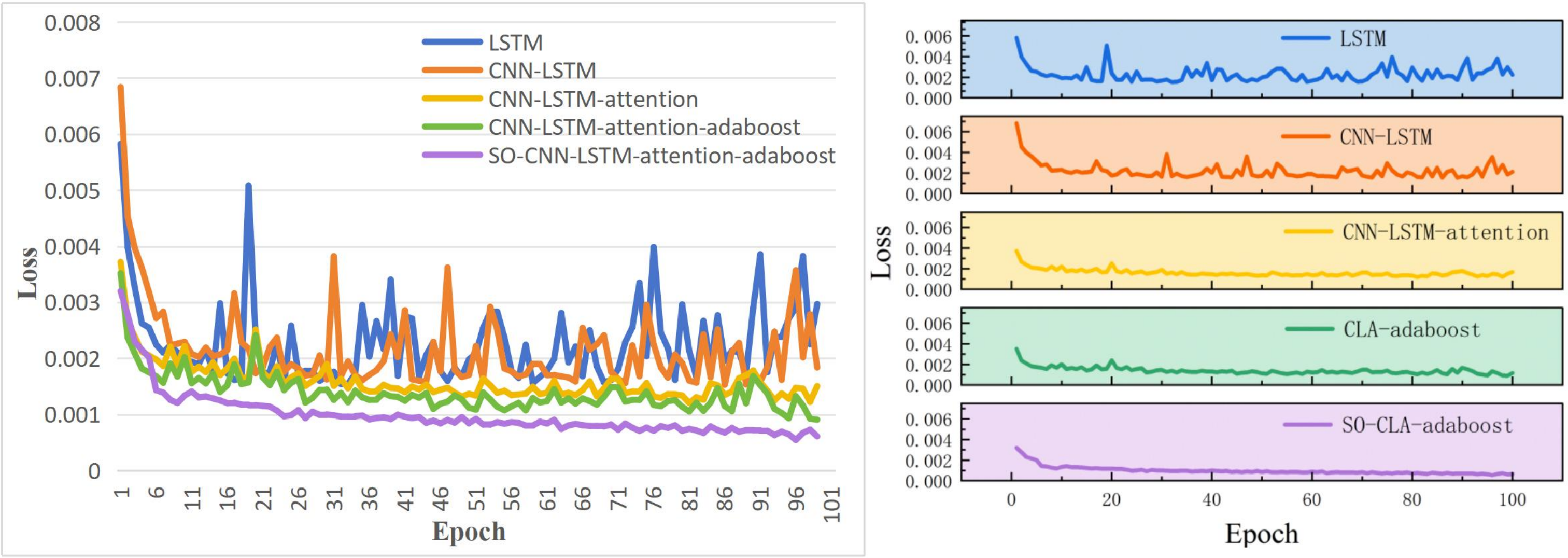

Figure15 Iterative plot of losses from ablation experiments

如图所示，使用 LSTM、CNN-LSTM、CNN-LSTM-attention、CNN-LSTM-attention-adaboost、SO-CNN-LSTM-attention-adaboost 五个模型进行训练时对应的 loss 值分别为 0.001542、0.001431、0.001202、0.000905、0.000544，加入对应模块后 loss 值的下降幅度分别为 7.14%、16.01%、24.71%、39.89%；达到 10%误差带所需的训练迭代次数分别为 18、31、50、96、94；达到最优 loss 值所需的训练迭代次数分别为 33、87、82、99、97。

Table7 Model convergence results

| | 稳定 10%误差带对应 epoch | 最优 loss 值对应 epoch | 最优 loss 值 | 对比 Loss 值 |
|---|---|---|---|---|

| | | | | 下降 |
|---|---|---|---|---|
| LSTM | 18 | 33 | 0.001542 | |
| CNN-LSTM | 31 | 87 | 0.001431 | 7.14% |
| CNN-LSTM-attention | 50 | 82 | 0.001202 | 16.01% |
| CNN-LSTM-attention-adaboost | 96 | 99 | 0.000905 | 24.71% |
| SO-CNN-LSTM-attention-adaboost | 94 | 97 | 0.000544 | 39.89% |

各模型在高度、经度、纬度三个维度上的预测结果与三维预测航迹分别如图和图所示：

在爬升阶段（0-95），各个模型均能捕捉到高度的上升趋势，但 LSTM 和 CNN-LSTM 的表现稍差，误差较为明显，而 CNN-LSTM-attention、CNN-LSTM-attention-adaboost 两大模型基本都能精准地预测高度的变化，在引入 SO 优化模块之后，表现地更加平滑。

在巡航阶段（95-270），LSTM 和 CNN-LSTM 模型在高度的预测上存在明显误差，无法准确地捕捉到高度的细微变化趋势，attention 模块和 adaboost 模块的加入显著提升了模型在巡航阶段的表现，将高度预测误差值降至 100m 以内，SO 优化模块的加入再次大幅度提升了预测性能，帮助其捕捉尽可能多的细微变化趋势，有效减少了巡航阶段的预测震荡，使模型表现更加平稳。

在降落阶段（270-310），LSTM 和 CNN-LSTM 模型在经度和纬度的预测上出现了剧烈的起伏震荡，在引入注意力机制后震荡幅度显著减小，经 adaboost 机制的多轮学习器择优后震荡逐渐消失，但距离真实值还是存在较明显的误差，SO 优化模块出色地处理好了经纬度连续值预测中的误差，展现了优越的拟合能力与稳定性。

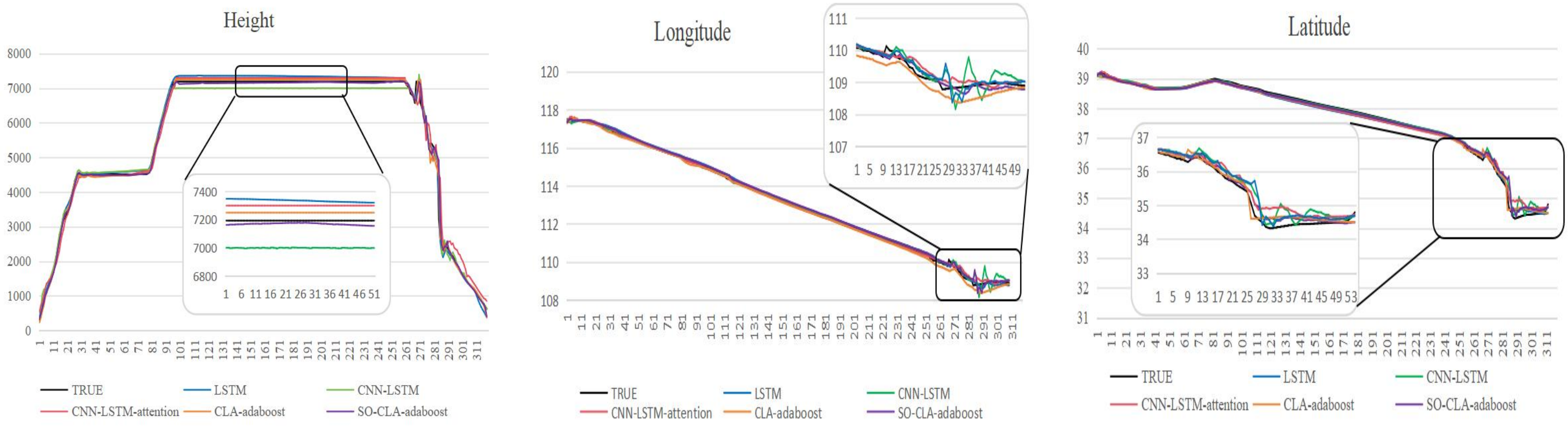

(a) (b) (c)

Figure16 Comparison of trajectory prediction in different dimensions

如图所示，LSTM 和 CNN-LSTM 模型的误差较大，RMSE 值分别为 526.4261、452.3430，420.2950、320.8560、125.4342，在依次引入 attention 机制和 adaboost 算法后，RMSE 值分别下降至 420.2950、320.8560，对比 CNN-LSTM 模型分别降低了 7.08%、29.20%，减少了预测迭代中的积累误差，特别是降低了高度预测中的波动与震荡，使得高度预测的精度显著提高；而改进后的 SO 优化模块的接入，不仅大幅度降低了预测误差，特别是在高度变化较为剧烈的起飞和下降阶段较迅速地捕捉到了浮动趋势，有效地弥补了现有模型对于高度骤升骤降反应过慢的缺陷，将最大绝对误差从 2136.5606 降至 609.4213，降幅高达 71.48%，极大地提升了模型在实际工程问题上的实用性。

Table8　Quantitative indicators of model prediction performance

| | RMSE | MAPE | MAE | MAXAE |
|---|---|---|---|---|
| LSTM | 526.4261 | 0.0536 | 203.5683 | 2308.2309 |
| CNN-LSTM | 452.3430 | 0.0478 | 180.6632 | 2234.3108 |
| CNN-LSTM-attention | 420.2950 | 0.0451 | 152.8691 | 2265.2707 |
| CLA-adaboost | 320.8560 | 0.0344 | 141.9650 | 2136.5606 |
| SO-CLA-adaboost | 125.4342 | 1.3504% | 51.9559 | 609.4213 |

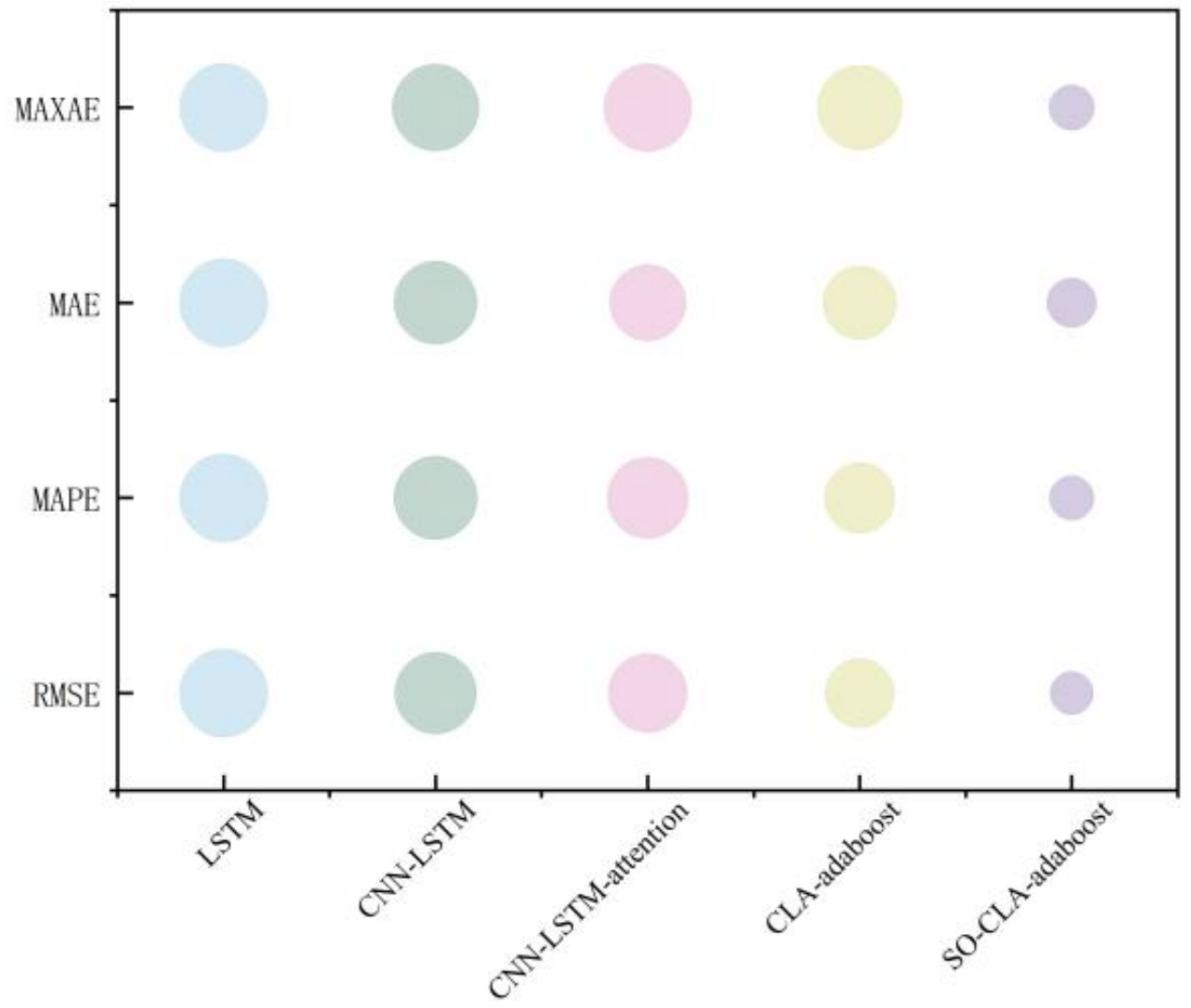

Figure17　Quantitative assessment analysis

# 五、结论

本研究针对中长期四维（4D）航迹预测具有较大局限性的问题，构建了基于多策略改进蛇群优化算法（SO）的 CNN-LSTM-attention-adaboost 混合神经网络模型，通过算法改进优化与模型集成实现了飞行器航迹预测精度的显著提升。主要研究结论如下：

（1）本文的 SO 算法利用佳点集初始化来提高种群初始质量，并在模拟蛇群行为模式和动态交互的过程中通过设置自适应化的阈值参数来保证种群演变的灵活性与可延续性，在探索阶段通过飞行游走机制强化全局勘探能力，在开发阶段使用“主要-辅助”双重变异策略进行深入开发，极大增强了算法的开发能力。本文还在 cec2022 测试集上对 SO 算法进行了全面的测试评估与考量，结果显示在几乎所有的测试函数上 SO 均展现出了远超其他算法性能的巨大优越性。

（2）本文的 SO-CLA-adaboost 模型可以精确地预测飞行器在飞行全程中的航迹，在天津至西安航线的 ADS-B 数据集上，SO-CLA-adaboost 模型的均方根误差（RMSE）和平均绝对百分比误差（MAPE）分别为 125.4342 和 1.3504%，较 LSTM 模型和 CNN-LSTM 模型分别降低了 76.17%和 72.27%，决定系数（$R^2$）更是达到了 0.9978，表明了模型对航迹数

据的高拟合度。

本文研究不足之处包括：（1）未考虑不同飞行器机型的飞行升降性能对航迹的影响；（2）未考虑极端天气如台风、暴雨等对航迹变更修改的影响。因此，在未来的研究中，飞行器自身性能等内在因素与极端天气等外部因素对于航迹的影响将是我的研究方向。